\documentclass[10pt,journal,compsoc]{IEEEtran}
\usepackage[utf8]{inputenc}
\usepackage{amsmath,amsfonts}
\usepackage{array}
\usepackage[caption=false,font=normalsize,labelfont=sf,textfont=sf]{subfig}
\usepackage{textcomp}
\usepackage{stfloats}
\usepackage{verbatim}
\usepackage{graphicx}
\usepackage{cite}
\usepackage{overpic}
\usepackage{amsfonts}
\usepackage{multirow}
\usepackage{amssymb}
\usepackage{float}
\usepackage{bm}
\usepackage{bbding}
\usepackage{pifont}
\usepackage{hyperref}
\usepackage{booktabs}
\usepackage{color}
\usepackage{colortbl}
\usepackage[table,xcdraw]{xcolor}
\usepackage[numbers,sort&compress]{natbib}
\definecolor{mygray}{gray}{.4}
\definecolor{rowgray}{gray}{.9}

\usepackage{algorithm}
\usepackage{algorithmic}
\begin{document}

\title{MAGIC: Meta-Ability Guided Interactive Chain-of-Distillation for Effective-and-Efficient Vision-and-Language Navigation}

\author{
% Anonymous Submission
\IEEEauthorblockN{Liuyi Wang, Zongtao He, Mengjiao Shen, Jingwei Yang, Chengju Liu, Qijun Chen,~\IEEEmembership{Senior Member,~IEEE}}

\thanks{This work was supported by the National Natural Science Foundation of
China under Grants (62233013). The authors are from Tongji University, Shanghai, China. (E-mails: \{wly, liuchengju, qjchen\}@tongji.edu.cn)}
\thanks{Corresponding authors: Qijun Chen, Chengju Liu}
% \thanks{Copyright \copyright 2022 IEEE. Personal use of this material is permitted. However, permission to use this material for any other purposes must be obtained from the IEEE by sending an email to pubs-permissions@ieee.org.}
}

% \date{May 2022}

\markboth{Journal of \LaTeX\ Class Files,~Vol.~14, No.~8, August~2015}%
{Shell \MakeLowercase{\textit{et al.}}: Bare Demo of IEEEtran.cls for Computer Society Journals}

% \IEEEpubid{0000--0000/00\$00.00~\copyright~2021 IEEE}
 
% =============
% Abstract
% =============
\IEEEtitleabstractindextext{
\begin{abstract}
Despite the remarkable developments of recent large models in Embodied Artificial Intelligence (E-AI), their integration into robotics is hampered by their excessive parameter sizes and computational demands. Towards the Vision-and-Language Navigation (VLN) task, a core task in E-AI, this paper reveals the great potential of using knowledge distillation for obtaining lightweight student models by proposing a Meta-Ability Guided Interactive Chain-of-distillation (MAGIC) method. Specifically, a Meta-Ability Knowledge Distillation (MAKD) framework is proposed for decoupling and refining the necessary meta-abilities of VLN agents. A Meta-Knowledge Randomization Weighting (MKRW) and a Meta-Knowledge Transferable Determination (MKTD) module are incorporated to dynamically adjust aggregation weights at the meta-ability and sample levels, respectively. Move beyond the traditional one-step unidirectional distillation, an Interactive Chain-of-Distillation (ICoD) learning strategy is proposed to allow students to give feedback to teachers, forming a new multi-step teacher-student co-evolution pipeline. 
Remarkably, on the R2R test unseen public leaderboard, our smallest model, MAGIC-S, with only 5\% (11M) of the teacher's size, outperforms all previous methods under the same training data. Additionally, our largest model, MAGIC-L, surpasses the previous state-of-the-art by 5.84\% in SPL and 3.18\% in SR. Furthermore, a new dataset was collected and annotated from our living environments, where MAGIC-S demonstrated superior performance and real-time efficiency. 
Our code is publicly available on \url{https://github.com/CrystalSixone/VLN-MAGIC}.
\end{abstract}

% =============
% Keywords
% =============
\begin{IEEEkeywords}
Vision-and-Language Navigation, Knowledge Distillation, Embodied AI, Vision-and-Language Reasoning
\end{IEEEkeywords}}

\maketitle

% =============
% 1-Introduction
% =============
\section{Introduction}
\label{sec_introduction}
The focus of Embodied Artificial Intelligence (E-AI) is pivotal for advancing the application of robots and agents in real-world environments. This paradigm shift necessitates models that exhibit excellent performance, efficiency, and generalization capabilities. Among the fundamental tasks that embody this trend, Vision-and-Language Navigation (VLN)~\cite{anderson2018vision} stands out as a canonical example. Given a natural language instruction like \textit{``Exit the kitchen, walk towards the hallway and stop at the top of the stairs,"} the agent must follow the instruction by exploring the visual environments and navigating to the target location automatically. However, current methods in VLN face significant challenges. The trend towards increasingly large models often results in redundant parameters and excessive computational complexity~\cite{wang2024causal,chen2022think,zhou2024navgpt}. This heightens the risk of overfitting and reduces the models' adaptability to edge devices like robots. Consequently, an important question arises: \textbf{\textit{Can we achieve embodied agents that balance performance and efficiency? }}
\begin{figure}[t]
    \centering
    \includegraphics[width=\linewidth]{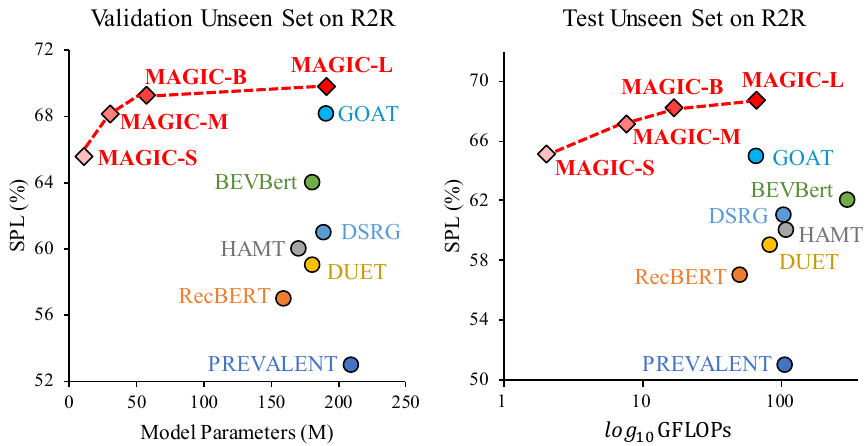}
    \caption{Model parameters and computational complexities versus accuracy comparison among state-of-the-art VLN methods.}  
    \label{fig_param_performance}
\end{figure}

Fortunately, knowledge distillation (KD)~\cite{hinton2015distilling} has emerged as a powerful technique for model compression, effectively transferring knowledge from larger, teacher models to smaller, student models. This approach not only reduces computational complexity but also maintains competitive performance. In the vision-and-language domain, KD has achieved some progress, demonstrating its ability to compress the model size while keeping the cross-modal representation capability~\cite{dai2022enabling,radford2021learning,kuang2023dlip,wu2023tinyclip}. However, applying KD to VLN tasks remains challenging since VLN is such a complex task that simultaneously involves single-modal feature extraction, cross-modal representation fusion, and long-term sequential reasoning with partially observable Markovian decision-making characteristics.

To effectively conduct KD in complex embodied reasoning models, addressing the fundamental questions of \textbf{\textit{what to distill}} and \textbf{\textit{how to distill}} is crucial. In response, we introduce the novel Meta-Ability Guided Interactive Chain-of-distillation (MAGIC) method, aiming to maximize efficiency by achieving high performance with minimal parameters. Our approach is grounded in the Meta-Ability Knowledge Distillation (MAKD) framework, inspired by the interdisciplinary nature of human learning, where different inductive knowledge is beneficial in different basic abilities, enabling more effective problem-solving strategies. For instance, mathematical knowledge enhances computational abilities, while physical knowledge improves understanding of fundamental laws of the universe. We apply this insight to realize KD for VLN. After carefully analyzing the current state-of-the-art (SoTA) VLN methods~\cite{wang2021causal,wang2023dual,chen2022think}, we propose decoupling the necessary meta-abilities into five components: visual perception, textual interpretation, local panoramic cross-modal matching, global topological cross-modal location, and behavioral decision-making. While the \textit{meta-ability} serves as an abstract conception, we further introduce the concept of \textit{``meta-knowledge"} to work as a concrete carrier for facilitating students to acquire the corresponding ability from their teachers.

Furthermore, we consider optimizing the potential imbalance problem of different meta-knowledge transfer losses, which could significantly impact learning performance and efficiency. The multi-task optimization methods typically sum losses directly or compute gradients for adaptive weighting~\cite{chen2018gradnorm,sener2018multi}. We propose Meta-Knowledge Randomization Weighting (MKRW), a simpler and more effective approach. By randomly assigning sampling weights to different meta-knowledge losses during the iterative process, MKRW effectively mitigates learning bias without increasing computational cost, enhancing the generalization and robustness of the knowledge distillation process.

Additionally, we recognize that the teacher model is not infallible, especially in the VLN domain where even the best-performing teacher models could also make incorrect predictions. Therefore, there is a significant risk of error propagation when the teacher guides the student through soft labels. To address this, we introduce the Meta-Knowledge Transferable Determination (MKTD) method, which leverages the teacher model's uncertainty to adjust the KD loss at a sample level, effectively reducing error interference from the teacher model and enhancing the reliability of meta-knowledge transfer.

Finally, we propose an Interactive Chain-of-Distillation (ICoD) learning strategy. This approach introduces a mutual assistant system, utilizing intermediate-sized models to bridge the information gap between the large teacher model and the small student model. Remarkably, this bridge is not just ``one-way" but ``interactive": not only does the teacher model guide the student, but as the student matures, it can also feedback and complement the teacher's knowledge. Surprisingly, this ICoD process not only enables us to obtain a small model with performance comparable to the large model but also further enhances the original large model's performance. We believe that these discoveries and proposals will significantly advance the rapidly growing E-AI community and have broad application prospects.

We validate the superiority of our proposed MAGIC method through experiments on various VLN datasets, including R2R~\cite{anderson2018vision}, RxR~\cite{ku2020room}, and a real-world dataset we collected. The results (Fig.~\ref{fig_param_performance}) show that, with the same in-domain training data, our smallest model, MAGIC-S, with only 5\% (11M) of the parameters, outperforms all previous state-of-the-art (SoTA) methods on the R2R test leaderboard. Our largest model, MAGIC-L, significantly improves the previous SoTA by 5.84\% in SPL and 3.18\% in SR, setting a new benchmark. Sim-to-real experiments with our real-world dataset further confirm our method's superiority, achieving better performance with reduced computational resources. Our code will be publicly available.

% =============
% 2-Introduction
% =============
\section{Related Work}
\label{sec_related_work}

\subsection{Vision-and-Language Navigation}
\label{subsec_vln}
Current Vision-and-Language Navigation (VLN)~\cite{anderson2018vision} methods predominantly revolve around three research directions: semantic encoding enhancement, historical dependency augmentation, and training strategy improvement.

(I) \textit{Semantic Encoding Enhancement} primarily concerns the semantic augmentation of both visual and linguistic modalities. For instance, Fried \textit{et al.}~\cite{fried2018speaker} introduced using panoramic images to represent visual environments. SOAT~\cite{moudgil2021soat} and DSRG~\cite{wang2023dual} integrated visual object features, and GeoVLN~\cite{huo2023geovln} and TAC~\cite{he2023learning} further incorporated depth information. FGR2R~\cite{hong2020sub}, MLANet~\cite{he2023mlanet} and GELA~\cite{cui2023grounded} proposed segmenting lengthy instructions into sub-instructions. NVEM~\cite{an2021neighbor} and OAAM~\cite{qi2020object} categorized instructions by semantics into different components. These methods have bolstered the agent's capacity for semantic understanding, underscoring the significance of visual and linguistic comprehension and encoding abilities.

(II) \textit{Historical Dependency Augmentation} addresses the requirement for agents to have a concrete representation of their current state and historical experiences. Early efforts utilized recurrent neural networks (\textit{e.g.,} LSTM) to update states~\cite{wang2020vision,an2021neighbor,dang2022unbiased,tan2024self}. VLN$\circlearrowright$BERT~\cite{hong2021vln} integrated the recurrent units on an encoder-only Transformer structure. HAMT~\cite{chen2021history} and HOP~\cite{qiao2023hop_plus} proposed treating historical memories as token sequences, engaging in the self-attention operations of memory modules. DUET~\cite{chen2022think} encoded visited nodes into a more structured graph form, enabling the agent's retrospection and global decision-making. DSRG~\cite{wang2023dual} combined explicit structured graph memory with implicit recurrent memory. Recently, some methods based on bird's-eye-view (BEV) scene graphs have also made good progress~\cite{an2022bevbert,wang2023gridmm}.

(III) \textit{Training Strategy Improvement} initially involves using reinforcement learning to boost the agent's online learning~\cite{wang2020vision}. Recent works have found that imitation learning is more effective and easier to converge~\cite{chen2022think,wang2024discovering}. 
Another direction is about data augmentation. Some methods proposed the speaker-follower framework~\cite{tan2019learning,wang2023pasts,wang2023res,wen2023vision} by constructing a separate instruction prediction model to expand the dataset. Some methods suggested collecting related navigation and video materials from a broader network~\cite{guhur2021airbert,lin2023learning}. ScaleVLN~\cite{wang2023scaling} introduced more environments from HM3D~\cite{ramakrishnan2021habitat} and Gibson datasets~\cite{xia2018gibson}. GOAT~\cite{wang2024causal} improved the model's generalizability by using causal learning to address dataset biases.
Recently, researchers have begun to explore the strong reasoning capabilities of large models to facilitate the VLN~\cite{zhou2024navgpt,yu2023l3mvn}.

Despite the remarkable achievements in this field, significant challenges remain in practical application deployment. One primary issue is the increasing complexity of models, which demands higher performance from edge devices, such as robots, resulting in substantial forward inference latency. This latency hinders the practical application of these models. While many studies solely focus on high accuracy, few address these practical difficulties. Therefore, this paper proposes a model compression method based on knowledge distillation, aiming to develop an effective and efficient VLN model suitable for physical deployment.

\subsection{Knowledge Distillation}
Knowledge Distillation (KD) is a model compression technique that facilitates the training of a smaller ``student" network under the guidance of a larger ``teacher" network. This method was initially introduced by Bucilua~\textit{et al.}~\cite{bucilua2006model} and later gained popularity through the work of Hinton~\textit{et al.}~\cite{hinton2015distilling}, who introduced a temperature scaling method to soften the output targets. 
Additionally, Romero~\textit{et al.}~\cite{adriana2015fitnets} introduced the concept of hint learning, utilizing the outputs from the teacher's hidden layers to supervise the student. 
Zagoruyko~\textit{et al.}~\cite{zagoruyko2016paying} highlighted the use of attention mechanisms as another effective strategy for knowledge transfer. Mirzadeh \textit{et al.}~\cite{mirzadeh2020improved} employed a teacher assistant to bridge the gap between the teacher and the student.

Studies have extensively explored knowledge distillation within single modalities, such as vision~\cite{liu2020structured,wu2022tinyvit} and language~\cite{sanh2019distilbert,gu2023minillm}. Recently, there has been a growing interest in cross-modal distillation. For example,
ALBEF~\cite{li2021align} proposed a momentum distillation approach that leverages pseudo-targets generated by a momentum model. DLIP~\cite{kuang2023dlip} comprehensively analyzed the distinct roles of visual and linguistic modes in KD. TinyCLIP~\cite{wu2023tinyclip} introduced techniques like affinity mimicking and weight inheritance to enhance the distillation for large-scale language-image pre-trained models. WSD~\cite{huang2023efficient} proposed a whitened similarity distillation method to distill cross-modal features. Guo \textit{et al.}~\cite{guo2022context} proposed a context-aware graph method to improve the comprehension of the semantic dependencies among implicit contexts.
It is important to note that these methods have primarily focused on straightforward single-stage classification tasks. In contrast, tasks in E-AI, such as VLN, demand a more complex, multi-stage prediction capability. Therefore, we dissect the required capabilities of embodied agents and propose a novel meta-ability guided knowledge distillation framework with the adaptive weighting strategy. Additionally, we move beyond the traditional single-step teacher-student pipeline by introducing an interactive chain-of-distillation learning approach to improve the performance of both teachers and students iteratively.
% =============
% 3-Preliminary
% =============
\section{Preliminary}
\label{sec_preliminary}

\subsection{Task Formulation for VLN}
\label{subsec_problem_def}
The VLN task involves training an embodied agent to navigate real indoor environments based on natural language instructions. The Matterport3D simulator~\cite{chang2017matterport3d} provides a graph-based environment $\mathcal{G}=\{\mathcal{E},\xi\}$, where $\mathcal{E}$ and $\xi$ represent navigable nodes and connectivity edges, respectively. The data is annotated as pairs of trajectory $\mathcal{T}=\{a_1, a_2, ..., a_N\}$ and instruction $\mathcal{I}=\{w_1,w_2,...,w_L\}$, where $a_i$ and $w_i$ present the visited nodes and the words, and $N$ and $L$ are their respective lengths. At the $t$-th step, the agent can observe a panoramic view $\mathcal{E}_t$ of its current location. This panoramic view is divided into three perspectives on the vertical view: upward looking ($+30^{\circ}$), level looking ($0^{\circ}$), and downward looking ($-30^{\circ}$). On the horizontal view, it is cut into $12$ sub-images at $60^{\circ}$ intervals, resulting in a total of $12\times 3=36$ sub-images. Each sub-image has a size of $640\times480$ pixels, and the vertical field of view is $60^{\circ}$. The orientation is represented as a 4D vector $\gamma = (\sin{\theta},\cos{\theta},\sin{\phi},\cos{\phi})$, where $\theta$ and $\phi$ denote the heading and elevation direction, respectively. During navigation, the agent needs to select the next point from visible candidates to follow the given instruction:
{\setlength\abovedisplayskip{3pt}
\setlength\belowdisplayskip{3pt}
\begin{align}
    P(a_1, a_2, ..., a_n|\mathcal{I}, \mathcal{G})=\prod_{i=1}^{N}P(a_i|a_1,...,a_{i-1}, \mathcal{I}, \mathcal{E}_i).
\end{align}}

The navigation is successful if the stop location falls within a certain threshold of the ground-truth position.

\begin{figure*}[thb]
    \centering
    \includegraphics[width=0.98\linewidth]{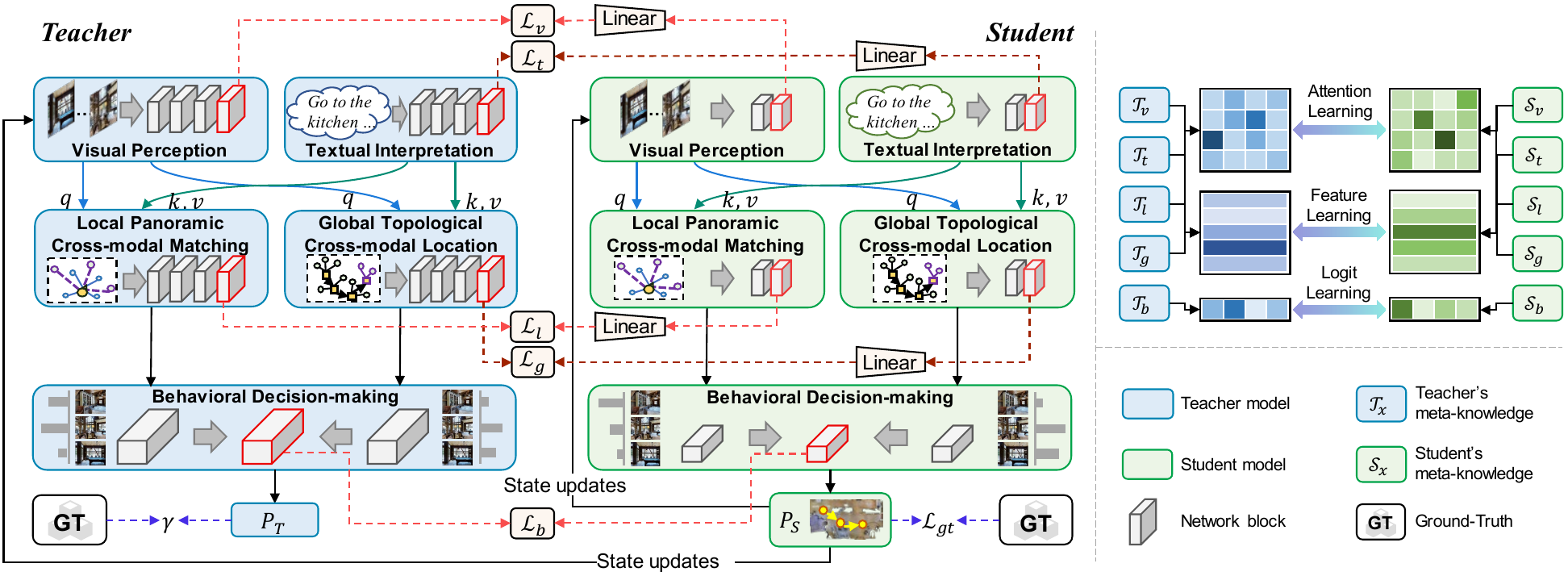}
    \caption{Architecture of the MAKD. Blue boxes denote the teacher model, while green boxes signify the student model. The MAKD framework identifies five core meta-abilities. The loss types of KD spanning attentions, features, and logits differ based on the targeted meta-ability.}  
    \label{fig_overall}
\end{figure*}

\subsection{Meta-Ability Framework for VLN}
\label{subsec_meta_ability_framework}
Due to the characteristics of VLN, including multimodal fusion and long-term sequential decision-making, current prevailing methods first employ vision-language multimodal encoders to encode both visual and linguistic inputs. 
Subsequently, a memory module, such as the topological graph, is utilized to store visited navigation paths structurally. Through the updating of the agent's observation state and memory units at each step, the navigation decision-making process is gradually completed. This architecture integrates multiple inductive biases. Through an in-depth analysis of previous SoTA methods~\cite{wang2024causal, wang2023dual, chen2022think, an2022bevbert}, we abstract and decouple it into the following five meta-abilities:

$\bullet$ \textit{Visual Perception Ability:} The agent should be able to fully leverage and perceive its visual environment. Typically, a pre-trained visual encoder (such as CLIP~\cite{radford2021learning}) is employed to extract raw image features. By connecting image features $V$ with their corresponding viewpoint offset angles $\gamma_v$, a learnable Transformer encoder is utilized for encoding and obtaining panoramic visual features $F_v$. Let $\phi$ and $\chi$ denote the linear layer and the transformer block:
{\setlength\abovedisplayskip{1pt}
\setlength\belowdisplayskip{1pt}
\begin{align}
    V &= \phi_v(\chi_{CLIP}(\mathcal{E})), \\
    F_v &= \chi_{self-attn}(V+\phi_{vp}(\gamma_v)).
\end{align}}

$\bullet$ \textit{Textual Interpretation Ability:} The agent should possess the ability to comprehend and interpret given long-text instructions. Similarly, we leverage a RoBERTa model~\cite{liu2019roberta} along with the absolute position $P_I$ to achieve end-to-end language parsing and encoding:
{\setlength\abovedisplayskip{1pt}
\setlength\belowdisplayskip{1pt}
\begin{align}
    F_I = \chi_{RoBERTa}(\phi_i(\mathcal{I})+\phi_{ip}(P_I)).
\end{align}}

$\bullet$ \textit{Local Panoramic Cross-modal Matching Ability:} 
Aligning semantic features from the visual panorama with guiding information contained in textual instructions is crucial for instruction following. The local visual panoramic sequences are combined with extra \texttt{[CLS]} and \texttt{[MEM]} tokens~\cite{wang2023dual}, denoted as $\widetilde{F}_v=\{\texttt{[CLS]},F_v,\texttt{[MEM]}\}$. We employ a cross-modal attention module~\cite{dou2022empirical} to update panoramic visual features (as $Q$) based on textual features (as $K$ and $V$):
{\setlength\abovedisplayskip{1pt}
\setlength\belowdisplayskip{1pt}
\begin{align}
    LC_v &= \phi_l(\widetilde{F}_v)+\phi_{lp}(\gamma_l), \\
    LC_c &= \chi_{co-attn}(LC_v, F_I, F_I).
\end{align}}

$\bullet$ \textit{Global Topological Cross-modal Location Ability:} It is essential for VLN to establish a topological map and perform global localization tracking autonomously. The Global Adaptive Aggregation (GAA) promotes the fusion of panoramas at the $t$-th step into the topological map~\cite{wang2023dual}. The cross-modal attention module is employed for integrating the map with the instructions at the global level:
{\setlength\abovedisplayskip{1pt}
\setlength\belowdisplayskip{1pt}
\begin{align}
    F_{g,t} &= GAA(F_{v,t}), \\
    \widetilde{F}_g &= \{\texttt{[CLS]},\{F_{g,t}\}_{t=1}^{T},\texttt{[MEM]}\}, \\
    GC_v &= \phi_g(\widetilde{F}_g) + \phi_{gp}(\gamma_g), \\
    GC_c &= \chi_{co-attn}(GC_v, \widetilde{F}_I, \widetilde{F}_I).
\end{align}}

$\bullet$ \textit{Behavioral Decision-making Ability:} This ability, closest to the output terminal, enables the agent to predict reliable and accurate behaviors during navigation. It leverages an adaptive $w$ that combines local and global fusion information, utilizing a prediction head $H(\cdot)$ (consisting of two linear layers and the intermediate nonlinear activated ReLU layer and layernorm layer) for behavioral output $b_n$:
{\setlength\abovedisplayskip{3pt}
\setlength\belowdisplayskip{3pt}
\begin{align}
    B_l &= H_l(LC_c), \, B_g = H_g(GC_c), \\
    B_f &= wB_l + (1-w)B_g, \\
    b_n &= \frac{\exp(B_{f,n})}{\sum_j \exp(B_{f,j})}.
\end{align}}
At each step, the agent prioritizes the candidate with the highest likelihood for its next move. If the \texttt{[CLS]} token is identified as the most likely option, the navigation is considered achieved and stopped.

% =============
% 4-Method
% =============
\section{Methodology}
\label{sec_method}
In this section, the details of the meta-ability guided interactive chain-of-distillation learning (MAGIC) are introduced. Initially, the meta-ability knowledge distillation (MAKD) strategy is described in Sec.~\ref{subsec_meta_ability_kd}. The meta-knowledge randomization weighting (MKRW) method is proposed in Sec.~\ref{subsec_MKRW}. The meta-knowledge transferable determination (MKTD) method, designed to mitigate noise interference, is presented in Sec.~\ref{subsec_mktd}. Finally, the interactive chain-of-distillation learning (ICoD) method is introduced in Sec.~\ref{subsec_mcdl}.

\subsection{Meta-Ability Knowledge Distillation}
\label{subsec_meta_ability_kd}
To explore the question of ``what to distill," we initially present the MAKD framework. As depicted in Fig.~\ref{fig_overall}, MAKD comprises a teacher model with a larger parameter count, and a student model with fewer parameters. Given the nature of the Transformer~\cite{vaswani2017attention}, the number of model parameters and computations can be approximated as:
{\setlength\abovedisplayskip{3pt}
\setlength\belowdisplayskip{3pt}
\begin{align}
    Param &= (12h^2+13h)l+Dh, \\
    FLOPs &= (24bsh^2+4bs^2h)l+2bshD,
\end{align}}
where $h$ is the dimension of hidden states, $b$ the batch size, $s$ the sequence length, $l$ the number of Transformer layers, and $D$ the vocabulary size. Consequently, by adjusting \(h\) and \(l\), we can efficiently scale down the student network.

The teacher model, through its comprehensive training, generates ability outputs rich in semantic information, thereby enhancing its final predictive. 
From the perspectives of representation learning and information theory, \textit{Mutual Information} (MI) measures the mutual dependence between the teacher ($T$) and the student ($S$). Maximizing the MI between the teacher and student models ensures that the student's representations closely mirror those of the teacher, enabling the student to inherit the teacher's superior capabilities. As delineated by~\cite{barber2004algorithm}, the MI of the teacher and student can be expressed as:
{\setlength\abovedisplayskip{3pt}
\setlength\belowdisplayskip{3pt}
\begin{align}
    I(T;S) &= H(T)-H(T|S) \nonumber \\
           &= -\mathbb{E}_T[\log p(T)] + \mathbb{E}_{T,S}[\log p(T|S)], \label{eq_mutual_1} 
\end{align}}
where $H(T)$ denotes the entropy of the teacher features, and $\mathbb{E}_{T,S}\log q(T|S)$ is the expected log probability that the student model accurately predicts the teacher's output. Since the true distribution $p(T|S)$ is unknown, we use $q(T|S)$ to estimate $p(T|S)$. Considering the non-negativity of the Kullback-Leibler (KL) divergence, we have:
{\setlength\abovedisplayskip{3pt}
\setlength\belowdisplayskip{3pt}
\begin{align}
    I(T;S) &= H(T)+\mathbb{E}_{T,S}[\log q(T|S)] \nonumber \\
    &\quad + \mathbb{E}_S[KL(p(T|S)||q(T|S))] \\
    &\geq H(T) + \mathbb{E}_{T,S}\log q(T|S). \label{eq_mutual_geq}
\end{align}}

In our MAKD framework, the subdivision of navigation capabilities into five distinct meta-abilities mirrors the disintegrability nature of human learning. 
By dividing complex tasks into manageable components, each meta-ability can address specific aspects of the problem, allowing for more effective collaboration and integration. To equip the model with the necessary meta-abilities, we further introduce the concept of \textit{``meta-knowledge."} Here, \textit{``meta-ability"} refers to the abstract manifestation, while \textit{``meta-knowledge"} acts as the tangible carrier. 
Let $\mathcal{T}=\{T_{v}, T_{t}, T_{l}, T_{g}, T_{b}\}$ and $\mathcal{S}=\{S_{v}, S_{t}, S_{l}, S_{g}, S_{b}\}$ denote the sets of the meta-knowledge (as discussed in Sec.~\ref{subsec_meta_ability_framework}) of the teacher and the student, respectively. Maximizing MI between these corresponding ability pairs inherently maximizes the MI between the teacher and student, as formalized below:
{\setlength\abovedisplayskip{3pt}
\setlength\belowdisplayskip{3pt}
\begin{align}
    \max \, I(\mathcal{T};\mathcal{S}) &= \max \sum_i I(T_i;S_i), \, i\in \{v,t,l,g,b\}. \label{eq_max_MI_system}
\end{align}}
Combined Eq.~(\ref{eq_max_MI_system}) with Eq.~(\ref{eq_mutual_geq}), we have:
{\setlength\abovedisplayskip{3pt}
\setlength\belowdisplayskip{3pt}
\begin{align}
    I(\mathcal{T};\mathcal{S}) \geq H(\mathcal{T}) + \sum_{i} \mathbb{E}_{T_i,S_i}\log q(T_i|S_i).
\end{align}}
Therefore, to improve the lower bound of the total MI, our objective becomes identifying the optimal network weights $\theta^{*}$ that maximize the sum of expectations, which can be reformulated to minimize the similarity losses:
{\setlength\abovedisplayskip{3pt}
\setlength\belowdisplayskip{3pt}
\begin{align}
    \min L_{kd} = -\sum_i L_i(T_i, S_i).
\end{align}}
\begin{figure}[t]
    \centering
    \includegraphics[width=0.98\linewidth]{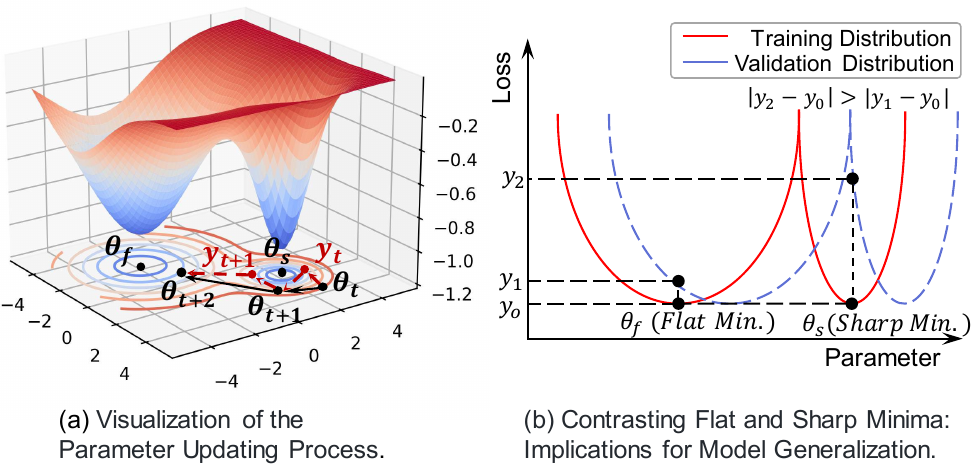}
    \caption{Illustration of parameter updating and effects of minimum type.}  
    \label{fig_minima}
\end{figure}
When using multiple loss functions to transfer meta-knowledge and optimize the network, additional noise is inevitably introduced. We hypothesize that the noise from each meta-knowledge module follows $\mathbb{E}(\omega_i^t)=0$ and that the variance $\mathbb{V}(\omega_i^t)$ is data-dependent. The aggregated noise is represented as \(\omega_t=\sum_i \omega_{i}^{t}\). We then examine how this additional noise affects the model's parameter updates.  
For simplicity, we disregard the inherent systematic noise arising from mini-batch sampling. As shown in Fig.~\ref{fig_minima}(a), it is common to encounter situations at step $t$ where the negative gradient for certain parameters $\theta_t$ leads to an unfavorable local minimum $\theta_s$ rather than the optimal solution $\theta_f$. We define $y_t \triangleq \theta_t-\eta \nabla L(\theta_t)$, where $\eta$ representing the learning rate. With the extra noise from multiple meta-knowledge, the actual update direction is modified to $\theta_{t+1}=\theta_t-\eta(\nabla L(\theta_t)+\omega_t)$. Inspired by~\cite{kleinberg2018alternative}, we use \(y_t\) as an analytical tool. Given that $\theta_{t+1} = y_t - \eta \omega_t$, we derive:
{\setlength\abovedisplayskip{3pt}
\setlength\belowdisplayskip{3pt}
\begin{align}
    y_{t+1} &= \theta_{t+1} - \eta \nabla L(\theta_{t+1}) \nonumber \\
    &= y_t-\eta \omega_t - \eta \nabla L(y_t-\eta \omega_t). \\
    \Rightarrow \mathbb{E}_{\omega_{t}}[y_{t+1}] &= y_t - \eta \nabla \mathbb{E}_{\omega_t}[L(y_t-\eta \omega_t)].
\end{align}}
Here, we define a new function $g_t$ as \(g_t(y)=\mathbb{E}_{\omega_t}[L(y-\eta\omega_t)]=\sum_{\omega_t}p(\omega_t)[f(y-\eta\omega_t)]\). This function \( g_t \) represents the original function \( L \) convolved with \(\eta\)-scaled gradient noise, resulting in a smoothed version of \( L \). The smoothing effect arises because the sharp local minima in \( L \) are short-range fluctuations in the function's landscape. The convolution operator averages these fluctuations over the scale defined by \(\eta\), thus smoothing out these sharp features. Fig.~\ref{fig_minima}(b) contrasts the effects of flat and sharp minima, with sharp minima being more sensitive to data bias, suggesting that flatter minima could enhance generalization~\cite{keskar2016large,zhang2021self}.

To implement the discussed strategies, as depicted in Fig.~\ref{fig_overall}, we establish pathways for knowledge transfer between the teacher and the student at the final layer of various meta-knowledge modules, as this layer is the most comprehensive and reliable source of each meta-knowledge. For transformer-based meta-knowledge modules (\(L_i\) where \(i \in \{v, t, l, g\}\)), the transfer includes both attentions (Eq.~\ref{eq_attn}) and features (Eq.~\ref{eq_feat}), using mean square error (MSE) as the loss function. For the behavioral decision-making module (\(L_b\)), the logits obtained from a softmax function, with a label smoothing temperature factor \(\tau\), are refined by the KL divergence loss (Eq.~\ref{eq_kd}).
{\setlength\abovedisplayskip{3pt}
\setlength\belowdisplayskip{3pt}
\begin{align}
    L_{attn} &= \frac{1}{NH} \sum_n \sum_h (AT_{T,h,n} - AT_{S,h,n})^2, \label{eq_attn} \\
    L_{feat} &= \frac{1}{N} \sum_n [FT_{T,n} - (FT_{S,n} W_s+b_s)]^2, \label{eq_feat} \\
    L_{logit} &= \frac{1}{N} \sum_n L_{KL}(\frac{T_b}{\tau} || \frac{S_b}{\tau})\tau^2, \label{eq_kd}
\end{align}}
where $AT$ and $FT$ mean the attention and features outputs, $W_{s} \in \mathbb{R}^{d_s \times d_t}$ and $b_s \in \mathbb{R}^{d_t}$ are learnable weights for transforming student features to match the teacher dimensions, $N$ denotes the batch size, and $H$ is the number of student's attention heads. The student's total loss is as follows:
{\setlength\abovedisplayskip{3pt}
\setlength\belowdisplayskip{3pt}
\begin{align}
    L_t = \frac{1}{N}\sum_n (\alpha L_{kd}(X_n)+(1-\alpha)L_{ce}(\hat{A}_n, P_S(X_n))),
\end{align}}
where \(\alpha\) balances the knowledge distillation loss \(L_{kd}\) and the cross-entropy loss \(L_{ce}\) of the ground-truth \(\hat{A}_n\) and the student's outputs \(P_S\) for the given input \(X_n\).

\subsection{Meta-Knowledge Randomization Weighting}
\label{subsec_MKRW}
Different meta-abilities suggest the presence of various inductive biases in an end-to-end learning system. Optimizing the concurrent learning of these biases is crucial for unlocking the full potential of MAKD. Therefore, inspired by recent advancements in multi-task learning~\cite{chen2018gradnorm,lin2021reasonable}, 
we introduce the Meta-Knowledge Randomization Weighting (MKRW) mechanism. Let $M$ denote the number of meta-abilities. The concept behind MKRW is both simple and potent: The process begins by randomly sampling a set of weights, denoted as \(\widetilde{\Lambda}=\{\widetilde{\lambda}_{i}\}_{i=1}^{M}\), from a predefined probability distribution, typically the standard normalized Gaussian distribution. These initial weights are subsequently normalized to be non-negative based on Eq.~\ref{eq_mkrw_lambda}, resulting in a new set of weights \(\Lambda=\{\lambda_i\}_i^{M}\):
{\setlength\abovedisplayskip{3pt}
\setlength\belowdisplayskip{3pt}
\begin{align}
\label{eq_mkrw_lambda}
    \lambda_i = \frac{K \exp(\widetilde{\lambda}_i/\tau)}{\sum_j \exp(\widetilde{\lambda}_j/\tau)},
\end{align}}
where $K$ and $\tau$ denote the scaling factor and the temperature, respectively. Given that \(\widetilde{\lambda}_i\) and \(\widetilde{\lambda}_j\) are i.i.d. random variables, it holds that for any \(1 \leq i, j \leq M\), the expected values of \(\lambda_i\) and \(\lambda_j\) are equal, \textit{i.e.}, \(\mathbb{E}(\lambda_i) = \mathbb{E}(\lambda_j)\). Denoting \(\lambda_i\) at the \(t\)-th step as \(\lambda^t_i\), we observe:
{\setlength\abovedisplayskip{3pt}
\setlength\belowdisplayskip{3pt}
\begin{align}
    \sum_{i=1}^{M}\mathbb{E}(\lambda_i)&=\sum_{i=1}^{M}\sum_{t=1}^{T}p(\lambda_{i}^{t})\lambda_{i}^{t}=\frac{\sum_{t=1}^T\sum_{i=1}^{M}\lambda_{i}^{t}}{T} \nonumber \\
    &=\frac{TK}{T}=K, \\
    \Rightarrow \mathbb{E}(\lambda_i)&=\frac{K}{M}.
\end{align}}
The parameter updating formula becomes:
{\setlength\abovedisplayskip{3pt}
\setlength\belowdisplayskip{3pt}
\begin{align}
    \theta_{t+1} = \theta_{t}-\eta\sum_i \lambda_{i}^{t}(\nabla L_i(\theta_t)+\omega_i^t).
\end{align}}
The iterative updating for parameters is as follows:
{\setlength\abovedisplayskip{3pt}
\setlength\belowdisplayskip{3pt}
\begin{align}
    y_t &\triangleq \theta_{t}-\eta \sum_i \lambda_{i}^{t} \nabla L_i(\theta_t), \\ 
    \theta_{t+1} &= y_t - \eta \sum_{i}\lambda_{i}^{t}\omega_{i}^{t}.
\end{align}}
Then, the subsequent update can be expressed as:
{\setlength\abovedisplayskip{3pt}
\setlength\belowdisplayskip{3pt}
\begin{align}
    &y_{t+1}=y_t-\eta \sum_i \lambda_{i}^{t}\omega_{i}^{t} - \eta \sum_{i}\lambda_{i}^{t} \nabla L_i (y_t-\eta \sum_i \lambda_{i}^{t}\omega_{i}^{t}), \\
    &\mathbb{E}_{\omega_{t}}[y_{t+1}] = y_t-\eta \sum_i \lambda_{i}^{t}\nabla \mathbb{E}_{\omega_{t}}[L_i(y_t-\eta\sum_i \lambda_{i}^{t}\omega_{i}^{t})].
\end{align}}
Let $f_{i}^{t}=\lambda_{i}^{t}\omega_{i}^{t}$. The variance can be expressed as:
{\setlength\abovedisplayskip{3pt}
\setlength\belowdisplayskip{3pt}
\begin{align}
    \mathbb{V}[f_{i}^{t}] &= \mathbb{V}[{\lambda_{i}^{t}}]\mathbb{V}[\omega_{i}^{t}]+\mathbb{V}[\lambda_{i}^{t}]\mathbb{E}[\omega_{i}^{t}]^2+\mathbb{V}[\omega_{i}^{t}]\mathbb{E}[\lambda_{i}^{t}]^2 \nonumber \\
    &= \mathbb{V}[{\lambda_{i}^{t}}]\mathbb{V}[\omega_{i}^{t}] + (\frac{K}{M})^2\mathbb{V}[\omega_{i}^{t}].
\end{align}}
This formulation reveals a critical insight: provided \(\frac{K}{M} \geq 1\), the variance \(\mathbb{V}[f_{i}^{t}]\) exceeds the original noise variance \(\mathbb{V}[\omega_{i}^{t}]\). According to~\cite{lin2021reasonable}, this property is significant as it demonstrates that the introduced randomness via MKRW enhances the model's ability to escape sharp local minima, facilitating better generalization. Compared to direct addition, manual adjustments, or gradient transformation weighting schemes, MKRW offers a simple yet effective dynamic weighting process for multiple meta-knowledge losses, improving the student's learning efficiency and performance.

\subsection{Meta-Knowledge Transferable Determination}
\label{subsec_mktd}
\begin{figure}[htb]
    \centering
    \includegraphics[width=0.9\linewidth]{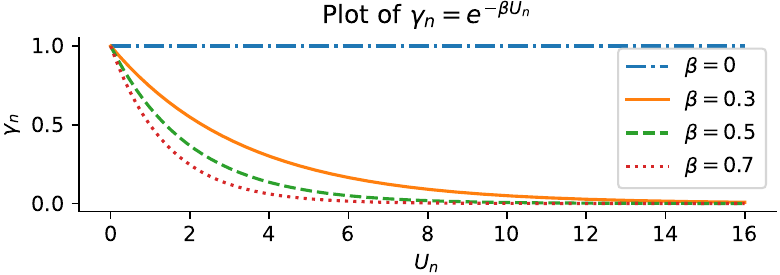}
    \caption{Impact of the attenuation coefficient $\beta$ with different values.}
    \label{fig_mktd}
\end{figure}

In the previous section, we explored adjusting weights at the meta-ability level. It's important to recognize that in KD, the teacher's predictions are not infallible; inaccuracies can occur, leading to potential misguidance when such knowledge is transferred to the student. To mitigate the adverse effects of transferring erroneous knowledge, we propose the Meta-Knowledge Transferable Determination (MKTD) strategy. This method assesses the transferability of knowledge from the teacher to the student, enhancing the process through adjustments to the weights at the sample level. For a given input \(X_n\) with its ground-truth action \(\hat{A}_n\) and the teacher's behavior prediction \(P_T(X_n)\) at the $t$-th step, we calculate the teacher's prediction uncertainty using their cross-entropy:
{\setlength\abovedisplayskip{3pt}
\setlength\belowdisplayskip{3pt}
\begin{align}
    U_n = -\hat{A}_n\log P_T(X_n),
\end{align}}
where a higher cross-entropy indicates a larger discrepancy between the teacher's prediction and the ground truth, signifying greater uncertainty in the teacher's knowledge. Eq.~\ref{eq_mktd} is used to calculate the transferable weights $\gamma_n$:
{\setlength\abovedisplayskip{3pt}
\setlength\belowdisplayskip{3pt}
\begin{align}
    \gamma_n = \exp(-\beta U_n), \label{eq_mktd}
\end{align}}
where \(\beta\) is the attenuation coefficient within the range of 0 to 1. As depicted in Fig.~\ref{fig_mktd}, a larger \(\beta\) results in a quicker decay in weight, indicating a stronger adjustment for more uncertain knowledge. The transferable weights $\gamma_n$ are assigned to all meta-ability losses with respect to the different samples. 

The MKTD complements the MKRW by adjusting weights for individual samples, in contrast to MKRW's meta-ability level adjustments. For instance, when employing the MSE loss to transfer feature meta-knowledge, the loss for the $n$-th sample is expressed as follows:
{\setlength\abovedisplayskip{3pt}
\setlength\belowdisplayskip{3pt}
\begin{align}
    L_{n}=\sum_i^M \lambda_i \exp(-\beta U_n)(T_n-S_n)^2,
\end{align}}
where \(\lambda_i\) is the $i$-th randomized ability weight from the MKRW. The gradient of this loss, which influences the update of the student's parameters $S_\theta$, is given by:
{\setlength\abovedisplayskip{3pt}
\setlength\belowdisplayskip{3pt}
\begin{align}
    \frac{\partial L_{n}}{\partial S_\theta} = 2\sum_i^M \lambda_i \exp (-\beta U_n)(S_n-T_n)\frac{\partial S_n}{\partial S_\theta}.
\end{align}}
This formulation ensures that samples with higher uncertainty exert less influence on the student's learning gradient, promoting a more reliable knowledge transfer.

\subsection{Interactive Chain-of-Distillation Learning}
\label{subsec_mcdl}
A general and effective training pipeline is essential to inspire the subsequent research on KD about ``how to distill" to achieve efficient and effective model training. As we aim to reduce the size of the student model, we observe that the disparity in knowledge capacity between the teacher (\textit{e.g.,} 100\% benchmark) and the student (\textit{e.g.,} reduced to 5\% benchmark) tends to widen, complicating the transfer of knowledge. Drawing from human learning experiences, such as the educational progression through high school, university, and graduate studies, we recognize that a phased learning system is more conducive to effective knowledge acquisition and skill development. Additionally, Unlike traditional KD processes that often ignore teacher feedback and optimization, we argue that the student can also provide valuable insights to supplement gaps in the teacher’s knowledge, fostering innovative thinking and enhancing the teacher’s capabilities. Based on these insights, we propose an Interactive Chain-of-Distillation (ICoD) pipeline, structured around four key steps:
\begin{figure}[t]
    \centering
    \includegraphics[width=0.9\linewidth]{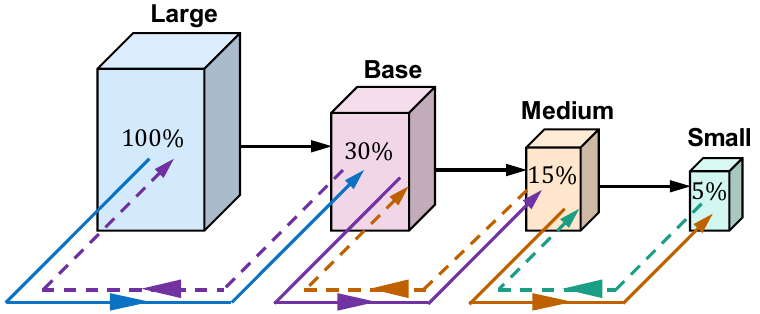}
    \caption{Illustration of the Interactive Chain-of-Distillation (ICoD).}
    \label{fig_ic}
\end{figure}

$\bullet$ \textbf{\textit{S1}}. 
Employ the commonly used training strategy to train a large teacher model. Since VLN is a multi-step decision process, the loss of each navigation is accumulated across steps. Let $P_T$ and $P_S$ denote the probabilities output by the teacher and the student models. We simplify the loss function using cross-entropy for each step, defined as:
{\setlength\abovedisplayskip{3pt}
\setlength\belowdisplayskip{3pt}
\begin{align}
    \mathcal{L}_{S1}=-\frac{1}{N} \sum_n \hat{A}_{n} \log P_T(X_{n}).
\end{align}}

$\bullet$ \textbf{\textit{S2}}. Train a smaller student model from scratch using the proposed MAKD framework, supported by MKRW $\lambda_i$ and MKTD $\gamma_n$ to dynamically adjust loss weights. During navigation, the teacher mirrors the student's actual actions to provide step-by-step guidance.
{\setlength\abovedisplayskip{3pt}
\setlength\belowdisplayskip{3pt}
\begin{align}
    \mathcal{L}_{S2}=-\frac{1}{N}\sum_n[&\alpha\sum_i^M\lambda_{i}\gamma_{n}L_{i}(P_{T,i}(X_n), P_{S,i}(X_n)) \nonumber \\
    &+(1-\alpha)\hat{A}_n \log P_S(X_n)].
\end{align}}

$\bullet$ \textbf{\textit{S3}}. Train the teacher and student models concurrently, with each alternating as the other's teacher. Given that both models have converged in previous steps, it is advisable to lower both the learning rate and the validation intervals.  
The respective losses for the teacher and student models are detailed as follows:
{\setlength\abovedisplayskip{3pt}
\setlength\belowdisplayskip{3pt}
\begin{align}
    \mathcal{L}_{S3}^T=-\frac{1}{N}\sum_n[&\alpha_t\sum_i^M\lambda_{i}\gamma_{S,n}L_{i}(P_{S,i}(X_n), P_{T,i}(X_n)) \nonumber \\
    &+(1-\alpha_t)\hat{A}_n \log P_T(X_n)], \\
    \mathcal{L}_{S3}^S=-\frac{1}{N}\sum_n[&\alpha_s\sum_i^M\lambda_{i}\gamma_{T,n}L_{i}(P_{T,i}(X_n), P_{S,i}(X_n)) \nonumber \\
    &+(1-\alpha_s)\hat{A}_n \log P_S(X_n)].
\end{align}}

$\bullet$ \textbf{\textit{S4}}. The refined student model is used as the new teacher to train a smaller model. As shown in Fig.~\ref{fig_ic}, this iterative process (\textit{S2} to \textit{S4}) is repeated until the balance between performance and efficiency meets specific requirements.

% =============
% 5-Experiment on sim
% =============
\section{Experiments on the Simulators}
\label{sec_experiment}
\begin{table*}[t]
\caption{Comparison with other SoTA methods on the R2R dataset~\cite{anderson2018vision}. `*': strong environments data augmentation with 1,291 extra scenes for training. Bold text indicates the best results with the original MP3D data, while blue text highlights important results.}
\label{tab:R2R_sota}
\resizebox{\linewidth}{!}{
\begin{tabular}{@{}l|c|c|cccc|cccc|cccc@{}}
\toprule
\multirow{2}{*}{Method} &
  \multirow{2}{*}{Param (M)$\downarrow$} &
  \multirow{2}{*}{GFLOPs$\downarrow$} &
  \multicolumn{4}{c|}{Validation Seen} &
  \multicolumn{4}{c|}{Validation Unseen} &
  \multicolumn{4}{c}{Test Unseen} \\
               &              &    & SR$\uparrow$    & SPL$\uparrow$   & NE$\downarrow$   & OSR$\uparrow$   & SR$\uparrow$    & SPL$\uparrow$   & NE$\downarrow$   & OSR$\uparrow$   & SR$\uparrow$    & SPL$\uparrow$   & NE$\downarrow$   & OSR$\uparrow$   \\ \midrule
$\text{ScaleVLN*}_{\text{\textcolor{mygray}{\,ICCV 2023}}}$~\cite{wang2023scaling}     & 181.02 & 82.13  & 80   & 75    & 2.12 & 87     & 79    & 70    & 2.34 & 87     & 77    & 68    & 2.73 & 83    \\ \midrule
$\text{PREVALENT}_{\text{\textcolor{mygray}{\,CVPR 2020}}}$~\cite{hao2020towards}     & 209.83 & 104.55 & 60    & 65    & 3.67 & -     & 57    & 53    & 4.73 & -     & 54    & 51    & 4.75 & 61    \\
$\text{RecBERT}_{\text{\textcolor{mygray}{\,CVPR 2021}}}$~\cite{hong2021vln}     & 159.99 & 50.38   & 72    & 68    & 2.90 & 79    & 63    & 57    & 3.93 & 69    & 63    & 57    & 4.09 & 70    \\
$\text{HAMT}_{\text{\textcolor{mygray}{\,NIPS 2021}}}$~\cite{chen2021history}       & 170.39 & 107.96 & 76    & 72    & 2.51 & 82    & 66    & 61    & 3.29 & 73    & 65    & 60    & 3.93 & 72    \\
$\text{ADAPT}_{\text{\textcolor{mygray}{\,CVPR 2022}}}$~\cite{lin2022adapt}       & 161.96 & 56.80  & 74    & 69    & 2.70 & -     & 66    & 59    & 3.66 & -     & 63    & 57    & 4.11 & -     \\
$\text{DUET}_{\text{\textcolor{mygray}{\,CVPR 2022}}}$~\cite{chen2022think}        & 181.02 & 82.13  & 79    & 73    & 2.28 & 86    & 72    & 60    & 3.31 & 81    & 69    & 59    & 3.65 & 76    \\
$\text{HOP+}_{\text{\textcolor{mygray}{\,TPAMI 2023}}}$~\cite{qiao2023hop_plus}  & 164.62 & 50.77   & 78    & 73    & 2.33 & -     & 67    & 61    & 3.49 & -     & 66    & 60    & 3.71 & -     \\
$\text{TD-STP}_{\text{\textcolor{mygray}{\,ACMMM 2023}}}$~\cite{zhao2022target}     & 171.87 & 130.63 & 77    & 73    & 2.34 & 83    & 70    & 63    & 3.22 & 76    & 67    & 61    & 3.73 & 72    \\
$\text{KERM}_{\text{\textcolor{mygray}{\,CVPR 2023}}}$~\cite{li2023kerm}       & 222.04 & 169.46  & 80    & 74    & 2.19 & -     & 72    & 61    & 3.22 & -     & 70    & 59    & 3.61 & -     \\ 
$\text{GeoVLN}_{\text{\textcolor{mygray}{\,CVPR 2023}}}$~\cite{huo2023geovln}      & 182.64 & 296.65  & 79    & 76    & 2.22 & -     & 68    & 63    & 3.35 & -     & 65    & 61    & 3.95 & -     \\ 
$\text{DSRG}_{\text{\textcolor{mygray}{\,IJCAI 2023}}}$~\cite{wang2023dual}       & 188.87 & 102.85 & 81    & 76    & 2.23 & 88    & 73    & 62    & 3.00 & 81    & 72    & 61    & 3.33 & 78    \\
$\text{BEVBert}_{\text{\textcolor{mygray}{\,ICCV 2023}}}$~\cite{an2022bevbert}     & 181.08 & 294.32  & 81    & 74    & 2.17 & 88    & 75    & 64    & 2.81 & 84    & 73    & 62    & 3.13 & 81    \\
$\text{GridMM}_{\text{\textcolor{mygray}{\,ICCV 2023}}}$~\cite{wang2023gridmm}      & 161.00 & 158.08 & 80    & 74    & 2.34 & 85    & 75    & 64    & 2.83 & -     & 73    & 62    & 3.35 & -     \\ 
$\text{GOAT}_{\text{\textcolor{mygray}{\,CVPR 2024}}}$~\cite{wang2024causal}        & 190.96 & 65.98  & 83.74 & \textbf{79.48} & 1.79 & 88.64 & 77.82 & 68.13 & 2.40 & 84.72 & 74.57 & 64.94 & 3.04 & 80.35 \\ \midrule
MAGIC-L (Ours)  & 190.96 & 65.98  & \textbf{83.84} & 79.33 & \textbf{1.73} & \textbf{88.93} & \textbf{78.88} & \textbf{69.82} & \textbf{2.22} & \textbf{86.12} & \textbf{77.02} & \textbf{68.73} & \textbf{2.75} & \textbf{81.88} \\
MAGIC-B (Ours)  & 57.88  & 16.69 & 81.78 & 76.80 & 2.00 & 86.29 & 77.01 & 69.26 & 2.48 & 84.08 & 75.13     & 68.25    & 2.99 & 79.10     \\
MAGIC-M (Ours)  & 30.23  & 7.52  & 82.08 & 77.19 & 1.84 & 87.17 & 77.39 & 68.16 & 2.52 & 85.10 & 75.53     & 67.12    & 2.85 & 81.09     \\
MAGIC-S (Ours)  & \textcolor{blue}{11.14}  & \textcolor{blue}{2.03}  & 78.16 & 71.10 & 2.37 & 85.41 & \textcolor{blue}{76.03} & \textcolor{blue}{65.07} & \textcolor{blue}{2.67} & \textcolor{blue}{84.89} & \textcolor{blue}{75.17}     & \textcolor{blue}{65.13}    & \textcolor{blue}{2.94} & \textcolor{blue}{81.88}     \\ \bottomrule
\end{tabular}}
\end{table*}

\subsection{Experimental Settings}

\subsubsection{Datasets}
The method is evaluated on two popular VLN datasets: Room-to-Room (R2R)~\cite{anderson2018vision} and Room-across-Room (RxR)~\cite{ku2020room}. The R2R dataset features real environments from 90 buildings, 7,189 paths, and 21,576 instructions averaging 29 words. It is divided into training, validation-seen (same buildings as in training), validation-unseen (different buildings from training), and test-unseen sets (assessed on an online platform for fair comparisons). The RxR dataset, an extension of R2R, addresses shortest path biases and includes more object references. We use the English segment of RxR, consisting of 42,002 instructions averaging 108 words. Additional experiments on our self-built dataset are introduced in Sec.~\ref{sec_exp_sim2real}.

\subsubsection{Metrics}
For R2R, we employ 4 metrics: \texttt{Navigation Error (NE)} calculates the distance between the predicted and actual stop locations. \texttt{Success Rate (SR)} gauges how often the predicted stop location is within a predefined distance from the true location. \texttt{Oracle Success Rate (OSR)} determines the frequency with which any point on the predicted path is within a certain distance of the goal. \texttt{Success Rate weighted by Inverse Path Length (SPL)} measures navigation effectiveness by combining success rate with the length of the route. For RxR, which does not use the shortest paths as ground truth, two additional metrics are introduced: \texttt{normalized Dynamic Time Warping (nDTW)}, assessing the alignment between the predicted and actual paths, and \texttt{success rate weighted by Dynamic Time Warping (sDTW)}, which evaluates both the accuracy of the predicted path and the path alignment.

\subsubsection{Implementation Details}
We adopt our previously proposed SoTA method, GOAT~\cite{wang2024causal}, as the benchmark teacher model. Following its experimental setup, both pre-training and fine-tuning training stages are included. CLIP ViT-B/16~\cite{radford2021learning} is used for image feature extraction. The loss balance $\alpha$ is set to $0.5$, the scaling factor $K$ and the temperature $\tau$ in MKRW to $5$ and $4$, and the attenuation coefficient $\beta$ in MKTD to $0.7$. The pre-training is conducted on two Tesla V100 GPUs, using a batch size of 96 for 200K iterations, with the AdamW optimizer and a learning rate of $5\times 10^{-5}$. The pre-training auxiliary tasks involve the masked language modeling (MLM)~\cite{devlin2018bert}, single action prediction (SAP)~\cite{chen2021history}, and cross-modal feature pooling (CFP)~\cite{wang2024causal} tasks. For data augmentation, we use PREVALENT~\cite{hao2020towards} for R2R and Marky-mT5-MP3D~\cite{wang2022less} for RxR. During the fine-tuning stage, in the individual training of student models (\textit{S2} in ICoD), we set the R2R batch size to 16 and the RxR batch size to 12, maintaining a learning rate of $5\times 10^{-5}$ for up to 100K iterations. During the joint training phase of both teacher and student models (\textit{S3} in ICoD), the learning rate is reduced to $5\times 10^{-6}$, with a maximum of 20,000 iterations. The causal learning modules in GOAT are used for feature debiasing during training. For R2R, a pre-trained speaker model~\cite{wang2023pasts} with environmental dropout (set at 0.5)~\cite{tan2019learning} is applied for back translation during fine-tuning. 
Model selection is based on performance on the validation unseen split, choosing the optimal checkpoint for R2R based on SR and SPL, and for RxR based on SR and sDTW.

\subsection{Comparisons with State-of-the-Arts}

\subsubsection{Results on R2R}
Tab.~\ref{tab:R2R_sota} compares the performance of our proposed MAGIC sets with different sizes (comparisons of network configurations are shown in Tab.~\ref{tab:model_size}) with other recent SoTA methods on the R2R dataset. We calculated the parameters (\texttt{Param}) and computation (\texttt{GFLOPs}) for all methods using their released codes. Computation was determined using the Python toolkit \texttt{thop}. For a fair comparison, we conducted single-step forward inference with a batch size of 8, instruction length of 44, candidates node of 3, and historical visited node of 6 across all methods. We only calculate the model's forward inference computational amount, excluding other data pre-processing and post-processing computations.

Specifically, ScaleVLN~\cite{wang2023scaling} generates extensive environments from the HM3D~\cite{ramakrishnan2021habitat} and Gibson~\cite{xia2018gibson} datasets, offering 150k $m^2$ of navigable areas and 1,291 scenes, which is significantly richer than MP3D (20k $m^2$, 61 scans). Despite attempting to train our model with this larger dataset, the possibly excessive noise resulted in slow training. Similar performance on unseen data and suboptimal performance on seen data are observed. Consequently, our experiments use only the original MP3D's data. RecBERT~\cite{hong2021vln}, ADAPT~\cite{lin2022adapt} and HOP+~\cite{qiao2023hop_plus} hold relatively low GFLOPs since they only use the images represented by candidate nodes for visual encoding rather than the panorama.

Without additional visual information (\textit{e.g.}, depth or semantics) or augmented scenes from other datasets, our experimental results demonstrate that, with much fewer parameters and GFLOPs, our model significantly surpasses previous methods. Notably, on the val-unseen set, our smallest model, MAGIC-S, relatively improves GridMM~\cite{wang2023gridmm} by 1.4\% in SR and 1.7\% in SPL, using only 6.9\% parameters and 3.1\% GFLOPs. Surprisingly, we also find that MAGIC-S shows higher performance than all previous methods based on original MP3D data, including the full GOAT model on the test unseen set (SR 75.17 \textit{vs.} 74.57, and SPL 65.13 \textit{vs.} 64.94). This significantly reveals the great generalization, effectiveness, and efficiency of our method. Furthermore, with only 4.5\% of the training scenes, our MAGIC-L model relatively outperforms ScaleVLN by 4.8\% in SR and 5.8\% in SPL on the val-seen set, achieving nearly consistent results in both the val-unseen and test-unseen sets. These results show that our method enables even very small student models to achieve high performance, while also further enhancing the original large teacher model without requiring a large amount of additional data. This is particularly important for model deployment on limited available datasets and real robots, which could significantly accelerate the adoption of advanced VLN models in various fields.

\begin{table*}[htb]
\centering
\caption{Network configurations for MAGIC sets.Noted that the results are without co-training in ICoD. $L$, $B$, $M$ and $S$ denote the large, base, medium, and small models. $\ell_{L},\,\ell_{P},\,\ell_{X}$: The number of layers of the textual language, visual panorama, and cross-modal encoders, respectively. $\mathcal{H}$: The dimensions of hidden features.}
\label{tab:model_size}
\begin{tabular}{l|llll|ll|cc|cc}
\toprule
 Model & $\ell_{L}$ & $\ell_{P}$ & $\ell_{X}$ & $\mathcal{H}$ & Param(M) (\%) & GFLOPs (\%) & Seen SR & Seen SPL & Unseen SR & Unseen SPL \\ \midrule
\#1 GOAT (MAGIC-L) & 6 & 2 & 3 & 768 & 190.96 (100) & 65.98 (100) & 83.74 & 79.48 & 77.82 & 68.13 \\ \midrule
\rowcolor{rowgray}
\#2 MAGIC-B & 6 & 2 & 3 & 384 & 57.88 (30.31) & 16.69 (25.30) & 81.39 & 76.67 & 76.33 & 67.86 \\
\#3 & 3 & 1 & 2 & 384 & 46.05 (24.11) & 10.46 (15.85) & 81.28 & 76.31 & 75.95 & 67.48 \\
\rowcolor{rowgray}
\#4 MAGIC-M & 6 & 2 & 3 & 256 & 30.23 (15.83) & 7.52 (11.40) & 79.04 & 74.35 & 75.78 & 66.65 \\
\#5 & 3 & 1 & 2 & 256 & 24.96 (13.07) & 4.75 (7.20) & 77.47 & 71.34 & 75.14 & 66.11 \\
\rowcolor{rowgray}
\#6 MAGIC-S & 6 & 2 & 3 & 128 & 11.14 (5.83) & 2.03 (3.08) & 76.59 & 71.20 & 73.44 & 65.56 \\ \bottomrule
\end{tabular}
\end{table*}

%=====
% Table of RxR
%=====
\begin{table}[]
\caption{Comparison on the RxR-English dataset~\cite{ku2020room}.}
\label{tab:rxr}
\centering
\large
\setlength\tabcolsep{3pt}
\resizebox{\linewidth}{!}{
\begin{tabular}{@{}l|cccc|cccc@{}}
\toprule
\multirow{2}{*}{Method} & \multicolumn{4}{c|}{Validation Seen} & \multicolumn{4}{c}{Validation Unseen} \\
               & SR$\uparrow$   & SPL$\uparrow$  & nDTW$\uparrow$ & sDTW$\uparrow$ & SR$\uparrow$   & SPL$\uparrow$  & nDTW$\uparrow$ & sDTW$\uparrow$ \\ \midrule
$\text{Baseline}_{\text{\textcolor{mygray}{\,EMNLP 2020}}}$~\cite{ku2020room}        & 28.6 & -    & 45.4 & 23.2 & 26.1 & -    & 42.4 & 21.0 \\
$\text{EnvDrop}_{\text{\textcolor{mygray}{\,NAACL 2019}}}$~\cite{tan2019learning}        & 48.1 & 44.0 & 57.0 & 40.0 & 38.5 & 34.0 & 51.0 & 32.0 \\
$\text{Syntax}_{\text{\textcolor{mygray}{\,NAACL 2021}}}$~\cite{li2021improving}         & 48.1 & 44.0 & 58.0 & 40.0 & 39.2 & 35.0 & 52.0 & 32.0 \\
$\text{EnvDrop+}_{\text{\textcolor{mygray}{\,ICLR 2021}}}$~\cite{shen2021much}       & -    & -    & -    & -    & 42.6 & -    & 55.7 & -    \\
$\text{SOAT}_{\text{\textcolor{mygray}{\,NIPS 2021}}}$~\cite{moudgil2021soat}           & -    & -    & -    & -    & 44.2 & -    & 54.8 & 36.4 \\
$\text{HOP+}_{\text{\textcolor{mygray}{\,TPAMI 2023}}}$~\cite{qiao2023hop_plus}           & 53.6 & 47.9 & 59.0 & 43.0 & 45.7 & 38.4 & 52.0 & 36.0 \\
$\text{FOAM}_{\text{\textcolor{mygray}{\,NAACL 2022}}}$~\cite{dou2022foam}           & -    & -    & -    & -    & 42.8 & 38.7 & 54.1 & 35.6 \\
$\text{ADAPT}_{\text{\textcolor{mygray}{\,CVPR 2022}}}$~\cite{lin2022adapt}       & 50.3 & 44.6 & 56.3 & 40.6 & 46.9 & 40.2 & 54.1 & 37.7 \\
$\text{CLEAR-C}_{\text{\textcolor{mygray}{\,NAACL 2022}}}$~\cite{li2022clear}     & -    & -    & -    & -    & 46.0 & 40.1 & 57.2 & 38.7 \\
$\text{PETL}_{\text{\textcolor{mygray}{\,ICCV 2023}}}$~\cite{qiao2023vln}       & 60.5 & 56.8 & 65.7 & 51.7 & 57.9 & 54.2 & 64.9 & 49.7 \\
$\text{GOAT}_{\text{\textcolor{mygray}{\,CVPR 2024}}}$~\cite{wang2024causal}           & \textbf{82.0} & 76.7 & 76.2 & 68.9 & 70.8 & 61.8 & 66.8 & 56.7 \\ \midrule
MAGIC-L (Ours)          & 81.3     & \textbf{77.5}     & \textbf{76.6}     & \textbf{69.2}    & \textbf{72.9}     & \textbf{65.4}     & \textbf{68.1}     & \textbf{58.7}     \\
MAGIC-B (Ours) & 75.3 & 69.9 & 71.5 & 62.3 & 70.4 & 62.9 & 66.5 & 56.4 \\
MAGIC-M (Ours) & 72.9 & 67.4 & 69.7 & 59.8 & 69.0 & 61.8 & 66.2 & 55.5 \\
MAGIC-S (Ours) & 70.2  & 63.8  & 65.7  & 56.0 & 68.1  & 59.8  & 64.0  & 53.6  \\ \bottomrule
\end{tabular}}
\end{table}

\subsubsection{Results on RxR} 
Tab.~\ref{tab:rxr} presents a comparison with other SoTA methods on the RxR-English dataset. The instructions and track lengths in the RxR dataset are significantly more complex than those in the R2R dataset, making RxR a considerably more challenging benchmark. 
It demonstrates that our MAGIC-L model demonstrates substantial performance improvements over the previous SoTA model GOAT. On the val-unseen subset, the SR, SPL, nDTW, and sDTW relatively improved by 5.3\%, 6.0\%, 1.5\%, and 3.9\%, respectively. Similar improvements can also be observed on the val-seen subset. As the model size decreases, we observe that the performance gap of MAGICs between the seen and unseen sets diminishes. Notably, MAGICs continue to exhibit strong performance on the unseen set, while showing a significant performance drop on the seen set. This phenomenon underscores the challenge posed by the RxR dataset, where instructions are exceedingly lengthy and complex. Understanding such ultra-complex instructions remains a critical hurdle for the VLN agent. Despite this, MAGICs still demonstrate a clear superiority in both performance and efficiency compared to previous methods.

\subsection{Quantitative Analysis}
\label{subsec:ablation}
This subsection presents detailed experimental results for each of our proposed modules. Unless otherwise specified, the ablation studies were conducted using the MAGIC-B model and reported on the R2R validation-unseen dataset.

\subsubsection{Impact of the meta-ability knowledge distillation (MAKD)}
\begin{figure}[t]
    \centering
    \includegraphics[width=1\linewidth]{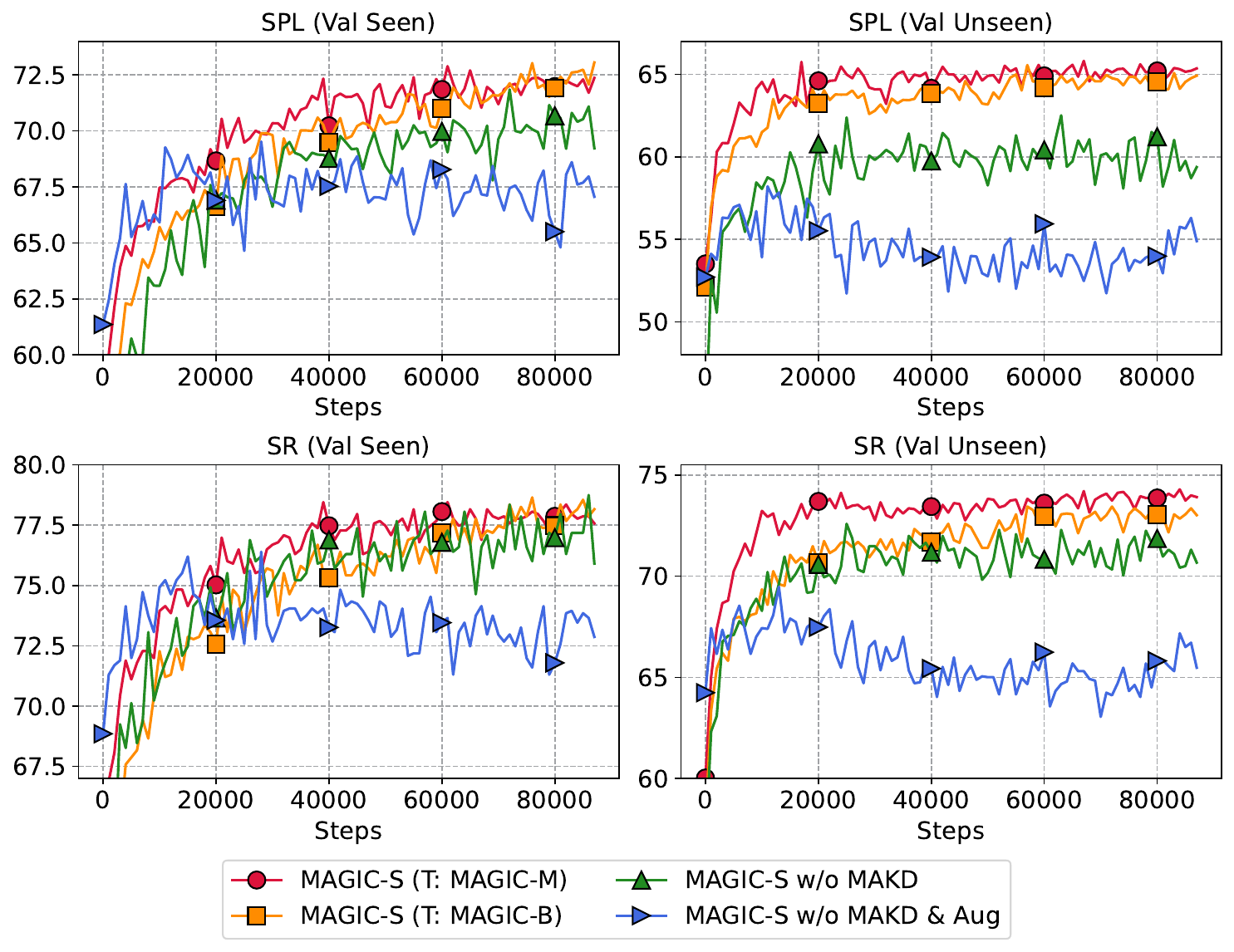}
    \caption{Training curves under different learning strategies. \textit{T} denotes the used teacher model, and \textit{Aug} means the usage of the speaker and causal inference.}
    \label{fig_learning_curve}
\end{figure}

As shown in Tab.~\ref{tab:MAKD_ablation}, when the KD process of all meta-abilities is eliminated ($\# 11$), that is, when the model relies only on the original supervised learning, performance is significantly lower than with MAKD training (SR $\downarrow 1.9\%$, and SPL $\downarrow 4.5\%$). When looking closely at the training curves, as shown in Fig.~\ref{fig_learning_curve}, the effect of MAKD on the generalization of small model learning is even more obvious. This demonstrates the need to enable MAKD when training the small models.

% ------
% Ablation study on MAKD.
% ------
\begin{table}[t]
\caption{Ablation Study of each meta-knowledge transfer in MAKD. An unticked meta-ability means that the corresponding meta-knowledge is not transferred for that ability.}
\label{tab:MAKD_ablation}
\centering
\begin{tabular}{l|lllll|cccc}
\toprule
Id & T & V & L& G & B & SR$\uparrow$ & SPL$\uparrow$ & NE$\downarrow$ & OSR$\uparrow$ \\ \midrule
\#1 & \checkmark & \checkmark & \checkmark & \checkmark & \checkmark & \textbf{76.33} & \textbf{67.86} & \textbf{2.44} & 83.61 \\ \midrule
\#2 & & \checkmark & \checkmark & \checkmark & \checkmark & 75.78 & 67.50 & 2.54 & \textbf{83.91} \\
\#3 & \checkmark &  & \checkmark & \checkmark & \checkmark & 76.20 & 67.21 & 2.52 & 83.64 \\
\#4 & \checkmark & \checkmark & & \checkmark & \checkmark & 75.95 & 66.58 & 2.64 & 83.27 \\
\#5 & \checkmark & \checkmark & \checkmark &  & \checkmark & 73.95 & 65.72 & 2.75 & 83.01 \\
\#6 & \checkmark & \checkmark & \checkmark & \checkmark &  & 76.19 & 66.87 & 2.53 & 84.89 \\ \midrule
\#7 & \checkmark &  &  &  &  & 74.93 & 64.83 & 2.68 & 83.06 \\
\#8 &  & \checkmark &  & &  & 73.61 & 64.16 & 2.85 & 81.86 \\
\#9 &  &  & \checkmark & \checkmark &  & 76.14 & 66.53 & 2.61 & 83.48 \\
\#10 &  &  &  &  & \checkmark & 73.82 & 65.71 & 2.84 & 81.44 \\ \midrule
\#11 &  &  & & &  & 74.88 & 64.82 & 2.74 & 83.61 \\ \bottomrule
\end{tabular}
\end{table}

\subsubsection{Impact of different meta-abilities in MAKD} 
As shown in Tab.~\ref{tab:MAKD_ablation} $\#2 \-- \#6$, when different individual meta-knowledge transfers are omitted, we find that the removal of the global topological cross-modal location ability (``G") causes the largest performance drop. This suggests that global memory knowledge is crucial for long-term decision-making in embodied agents. Eliminating behavior decision-making (``B") slightly reduces performance but increases the OSR value, indicating that without behavior knowledge transfer, the student model becomes more hesitant and unsure about when to stop. Tab.~\ref{tab:MAKD_ablation} $\#7 \-- \#10$ shows the comparison of adding individual meta-knowledge transferring. The results show that when the transfer of meta-knowledge is incomplete, the learning performance is not stable, and sometimes even worse. This indicates that the meta-ability of the agent is complementary to each other, and a richer and more complete meta-knowledge should be transmitted as much as possible, so that the agent can master more comprehensive meta-abilities.

% ----
% Figure of learning curves.
% ----

\subsubsection{Impact of different features on KD}
\begin{figure}[t]
    \centering
    \includegraphics[width=\linewidth]{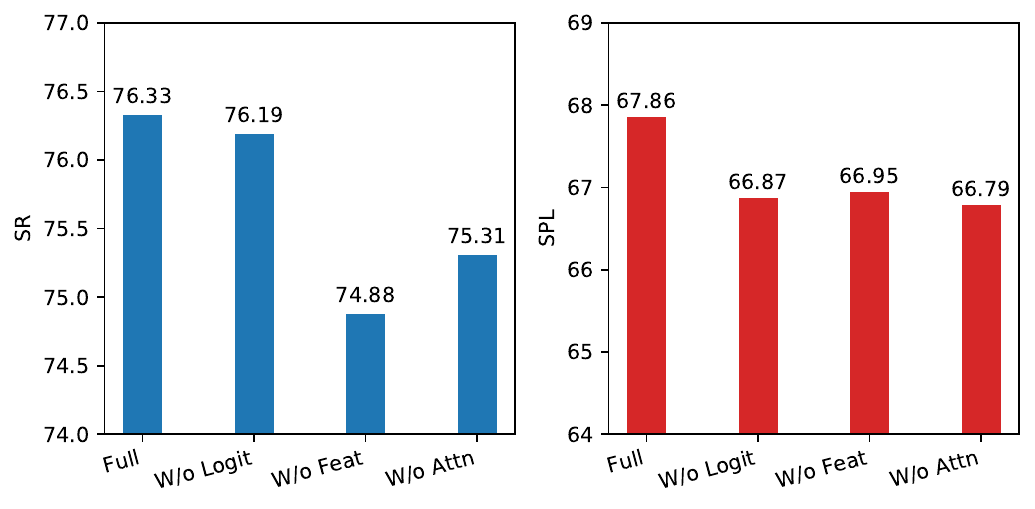}
    \caption{Comparison of different features involved in MAKD.}
    \label{fig_kd_type}
\end{figure}

Since all other meta-knowledge modules except the behavior prediction module are Transformer layers, we transfer their hidden layer features and attention features at the same time. Fig.~\ref{fig_kd_type} shows that the ablation of feature, attention, and logits will all lead to the degradation of training performance. When the knowledge transfer of the hidden feature is eliminated, the performance of SR decreases the most (SR $\downarrow 1.9\%$).

% ----
% Figure of KD types.
% ----

\subsubsection{Impact of the meta-knowledge randomization weighting (MKRW)}
\begin{figure}[t]
    \centering
    \includegraphics[width=\linewidth]{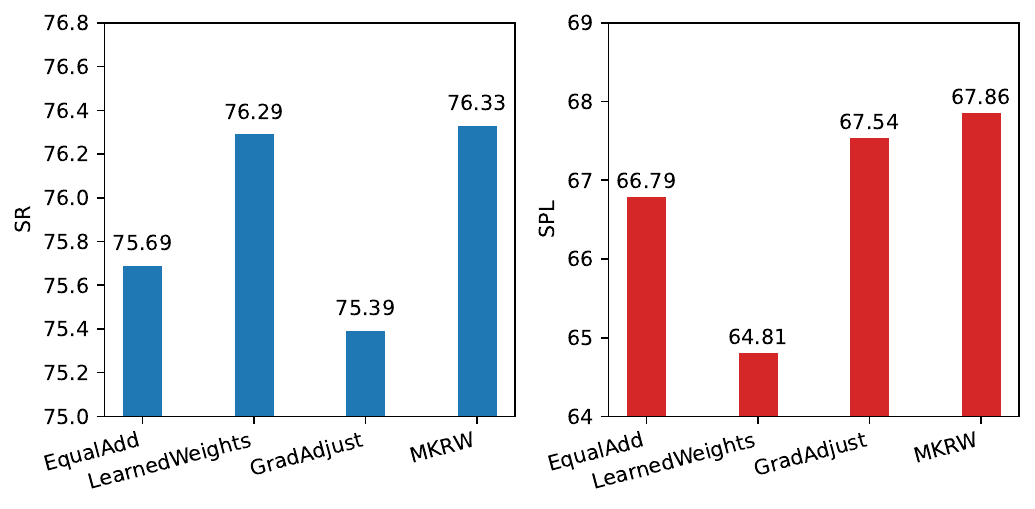}
    \caption{Comparison of different weighting methods with MKRW.}
    \label{fig_mkrw}
\end{figure}

Fig.~\ref{fig_mkrw} compares several different weighting methods. \texttt{EqualAdd} means that adding the different meta-knowledge losses directly. This is the most likely combination strategy to come up with. \texttt{LearnedWeights} means that the learnable weight $W\in \mathbb{R}^{M\times 1}$ is initialized randomly for $M$ kinds of meta-knowledge losses, and its value is adjusted adaptively during the learning process. \texttt{GradAdjust} is the strategy in which the weight of the corresponding loss is adjusted according to the gradient of the last layer of different meta-knowledge modules. The greater the gradient, the smaller the weight, to balance the losses. The final \texttt{MKRW} is our proposed weighting method discussed in Sec.~\ref{subsec_MKRW}. Without any parameters that need to be trained or cumulative calculation, our MKRW achieves both the highest SR and SPL across the other three kinds of weighting strategies, showing its simplicity and effectiveness.

% ----
% Figure of different weighting methods in MKRW.
% ----

\subsubsection{Impact of the meta-knowledge teacher determination (MKTD)}
\begin{figure}[thb]
    \centering
    \includegraphics[width=\linewidth]{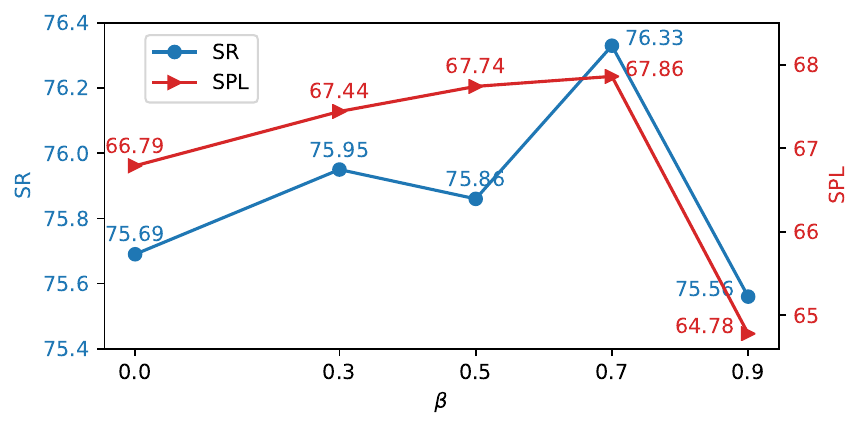}
    \caption{Comparison of different values of $\beta$ in MKTD. Greater $\beta$ means greater attenuation.}
    \label{fig_mktd_beta}
\end{figure}

As shown in Fig.~\ref{fig_mktd_beta}, when no MKTD is used ($\beta=0$), the performance gets significantly drop (SPL $\downarrow 1.6\%$). This supports our conjecture that the teacher model may produce additional noise that affects student learning. The results achieve the optimal when $\beta$ is set to $0.7$, which means that it is effective to adjust the weight of knowledge transfer on different samples dynamically based on the uncertainty of teacher model output. Additionally, when the attenuation factor is too large ($\beta=0.9$), the SPL reduces by 4.5\%, which indicates that excessive suppression of the distillation inhibits the advantage of knowledge transfer.

% ----
% Table of IC
% ----
\begin{table*}[t]
\caption{Performance of the interactive learning step (S3) in ICoD on R2R and RxR. The raised values have been bolded.}
\label{tab:IC}
\centering
\begin{tabular}{@{}lcccccccccccccc@{}}
\toprule
\multicolumn{1}{l|}{\multirow{2}{*}{Method}} & \multicolumn{3}{c|}{R2R Validation Seen} & \multicolumn{3}{c||}{R2R Validation Unseen} & \multicolumn{4}{c|}{RxR Validation Seen} & \multicolumn{4}{c}{RxR Validation  Unseen} \\
\multicolumn{1}{l|}{} & SR$\uparrow$ & SPL$\uparrow$ & \multicolumn{1}{c|}{OSR$\uparrow$} & SR$\uparrow$ & SPL$\uparrow$ & \multicolumn{1}{c||}{OSR$\uparrow$} & SR$\uparrow$ & SPL$\uparrow$ & NDTW$\uparrow$ & \multicolumn{1}{c|}{SDTW$\uparrow$} & SR$\uparrow$ & SPL$\uparrow$ & NDTW$\uparrow$ & SDTW$\uparrow$ \\ \midrule
\rowcolor{rowgray}
\multicolumn{15}{c}{\textit{Teacher: MAGIC-L \& Student: MAGIC-B}} \\ \midrule
\multicolumn{1}{l|}{L-before} & 83.74 & 79.48 & \multicolumn{1}{c|}{88.64} & 77.82 & 68.13 & \multicolumn{1}{c||}{84.72} & 81.97 & 76.72 & 76.16 & \multicolumn{1}{c|}{68.85} & 70.75 & 61.81 & 66.84 & 56.72 \\
\multicolumn{1}{l|}{L-after} & \textbf{83.84} & 79.37 & \multicolumn{1}{c|}{\textbf{88.93}} & \textbf{78.76} & \textbf{69.78} & \multicolumn{1}{c||}{\textbf{85.99}} & 81.29 & \textbf{77.46} & \textbf{76.62} & \multicolumn{1}{c|}{\textbf{69.16}} & \textbf{72.89} & \textbf{65.41} & \textbf{68.12} & \textbf{58.69} \\ \midrule
\multicolumn{1}{l|}{B-before} & 81.39 & 76.67 & \multicolumn{1}{c|}{86.09} & 76.33 & 67.86 & \multicolumn{1}{c||}{83.61} & 72.51 & 67.37 & 70.92 & \multicolumn{1}{c|}{61.07} & 67.37 & 60.06 & 66.12 & 55.02 \\
\multicolumn{1}{l|}{B-after} & \textbf{81.59} & \textbf{77.50} & \multicolumn{1}{c|}{\textbf{86.78}} & \textbf{76.86} & \textbf{68.45} & \multicolumn{1}{c||}{\textbf{84.42}} & \textbf{75.57} & \textbf{68.55} & \textbf{71.30} & \multicolumn{1}{c|}{\textbf{61.15}} & \textbf{69.14} & \textbf{62.63} & \textbf{66.45} & \textbf{56.24} \\ \midrule
\rowcolor{rowgray}
\multicolumn{15}{c}{\textit{Teacher: MAGIC-B \& Student: MAGIC-M}} \\ \midrule
\multicolumn{1}{l|}{B-before} & 81.59 & 77.50 & \multicolumn{1}{c|}{86.78} & 76.86 & 68.45 & \multicolumn{1}{c||}{84.42} & 75.57 & 68.55 & 71.30 & \multicolumn{1}{c|}{61.15} & 69.14 & 62.63 & 66.45 & 56.24 \\
\multicolumn{1}{l|}{B-after} & \textbf{81.78} & 76.80 & \multicolumn{1}{c|}{86.29} & \textbf{77.01} & \textbf{69.26} & \multicolumn{1}{c||}{84.08} & 75.26 & \textbf{69.85} & \textbf{71.50} & \multicolumn{1}{c|}{\textbf{62.34}} & \textbf{70.40} & \textbf{62.87} & \textbf{66.49} & \textbf{56.42} \\ \midrule
\multicolumn{1}{l|}{M-before} & 79.04 & 74.35 & \multicolumn{1}{c|}{85.01} & 75.78 & 66.65 & \multicolumn{1}{c||}{83.57} & 70.54 & 64.29 & 68.06 & \multicolumn{1}{c|}{58.00} & 65.99 & 59.12 & 64.91 & 53.82 \\
\multicolumn{1}{l|}{M-after} & \textbf{80.61} & \textbf{76.44} & \multicolumn{1}{c|}{\textbf{86.09}} & \textbf{76.50} & \textbf{67.65} & \multicolumn{1}{c||}{\textbf{84.16}} & \textbf{73.43} & \textbf{67.83} & \textbf{69.39} & \multicolumn{1}{c|}{\textbf{60.11}} & \textbf{68.51} & \textbf{60.59} & 64.84 & \textbf{54.36} \\ \midrule
\rowcolor{rowgray}
\multicolumn{15}{c}{\textit{Teacher: MAGIC-M \& Student: MAGIC-S}} \\ \midrule
\multicolumn{1}{l|}{M-before} & 80.61 & 76.44 & \multicolumn{1}{c|}{86.09} & 76.50 & 67.65 & \multicolumn{1}{c||}{84.16} & 73.43 & 67.83 & 69.39 & \multicolumn{1}{c|}{60.11} & 68.51 & 60.59 & 64.84 & 54.36 \\
\multicolumn{1}{l|}{M-after} & \textbf{82.08} & \textbf{77.19} & \multicolumn{1}{c|}{\textbf{87.17}} & \textbf{77.39} & \textbf{68.16} & \multicolumn{1}{c||}{\textbf{85.10}} & 72.85 & 67.41 & \textbf{69.70} & \multicolumn{1}{c|}{59.80} & \textbf{69.02} & \textbf{61.84} & \textbf{66.16} & \textbf{55.48} \\ \midrule
\multicolumn{1}{l|}{S-before} & 76.59 & 71.20 & \multicolumn{1}{c|}{81.29} & 73.44 & 65.56 & \multicolumn{1}{c||}{80.20} & 69.45 & 62.46 & 63.21 & \multicolumn{1}{c|}{52.48} & 66.60 & 58.12 & 63.02 & 52.82 \\
\multicolumn{1}{l|}{S-after} & \textbf{78.16} & 71.10 & \multicolumn{1}{c|}{\textbf{85.41}} & \textbf{76.03} & 65.07 & \multicolumn{1}{c||}{\textbf{84.89}} & \textbf{70.23} & \textbf{63.79} & \textbf{65.74} & \multicolumn{1}{c|}{\textbf{55.98}} & \textbf{68.09} & \textbf{59.78} & \textbf{63.96} & \textbf{53.60} \\ \bottomrule
\end{tabular}
\end{table*}

% ----
% Figure of beta in MKTD.
% ----

\subsubsection{Impact of the interactive chain-of-distillation (ICoD)}
The basis of ICoD is similar to the assistant method that uses a medium-sized model to help knowledge transfer~\cite{mirzadeh2020improved}. The learning curves in Fig.~\ref{fig_learning_curve} show that using MAGIC-M as a teacher to train MAGIC-S can get better results than using MAGIC-B as a teacher. This trend is consistent across other model pairs, indicating the effectiveness of step-by-step training via a chain of knowledge distillation.

\begin{figure}[t]
    \centering
    \includegraphics[width=\linewidth]{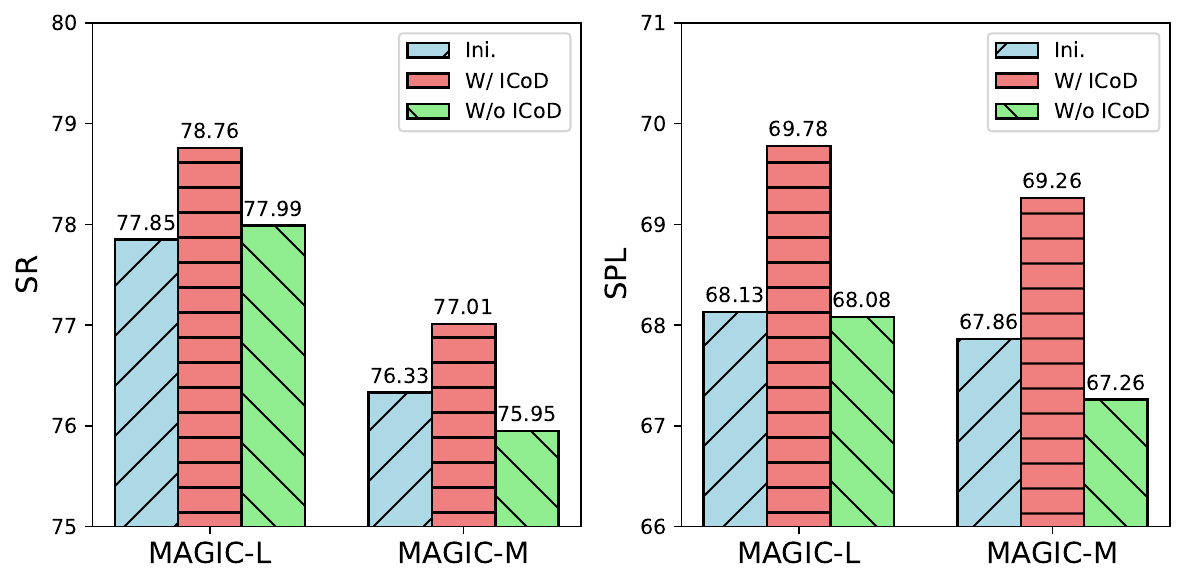}
    \caption{Ablation study of ICoD. The \textit{Ini.} models were continually trained with (\textit{W/ ICoD}) and without (\textit{W/o ICoD}) the student's feedback.}
    \label{fig_ic_ablation}
\end{figure}

Another core aspect of ICoD is the interactive teaching between the student and teacher models, providing supplementary knowledge to each other. Tab.~\ref{tab:IC} compares performance before and after the co-training. It clearly shows that with the help of ICoD, most metrics on both datasets get significant improvements. Importantly, ICoD enhances not only the small student model but also the large teacher model. To verify that improvements stem from mutual knowledge exchange rather than continuous learning, we conducted ablation experiments in Fig.~\ref{fig_ic_ablation}. It proves that without the student's feedback, the teacher's continuous training led to severe overfitting. 
% ----
% Figure of ablation of IC.
% ----

\subsection{Qualitative Analysis}
\label{subsec:visualization}

\subsubsection{Visualization of Navigation Trajectories}
\begin{figure*}[thb]
    \centering
    \includegraphics[width=0.98\linewidth]{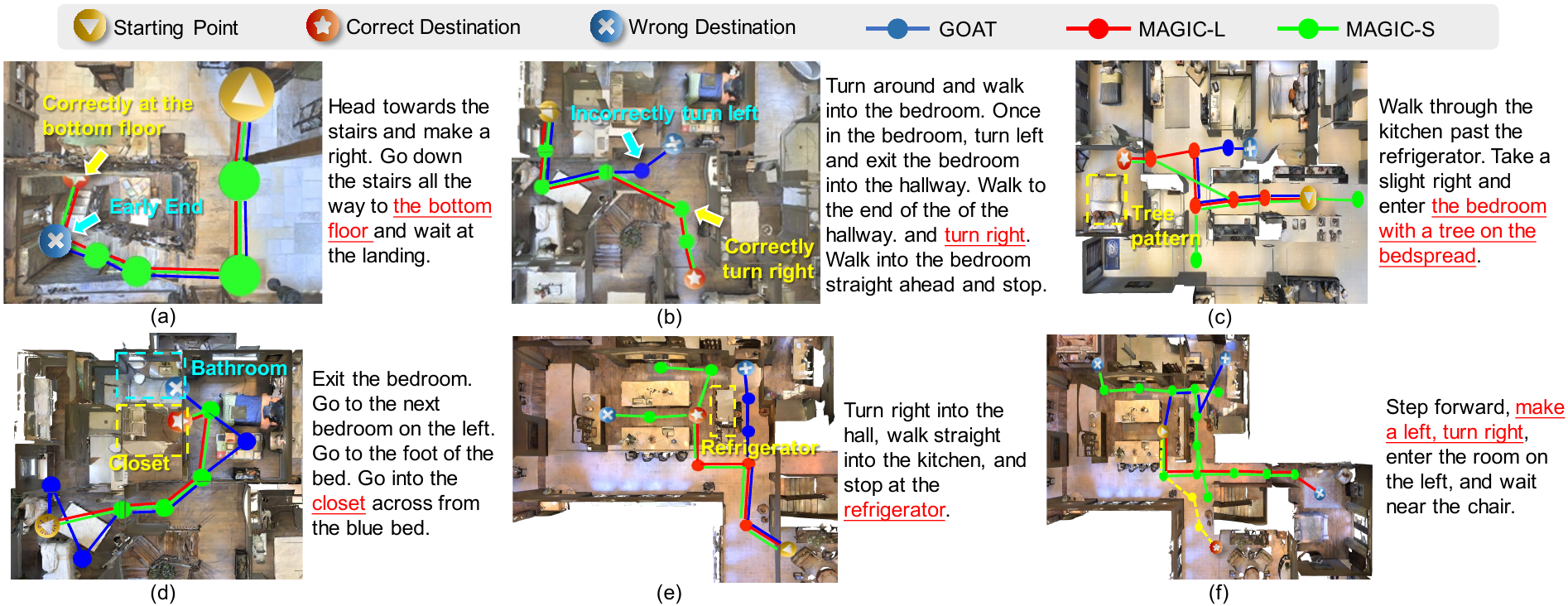}
    \caption{Visualized predicted trajectories are compared on the R2R validation unseen set. The given instructions are shown on the right of each figure, with the key reference text in red. Key navigation nodes are emphasized in the diagrams with specially colored text to aid comprehension.}
    \label{fig_vis_paths}
\end{figure*}

\begin{figure}[thb]
    \centering
    \includegraphics[width=\linewidth]{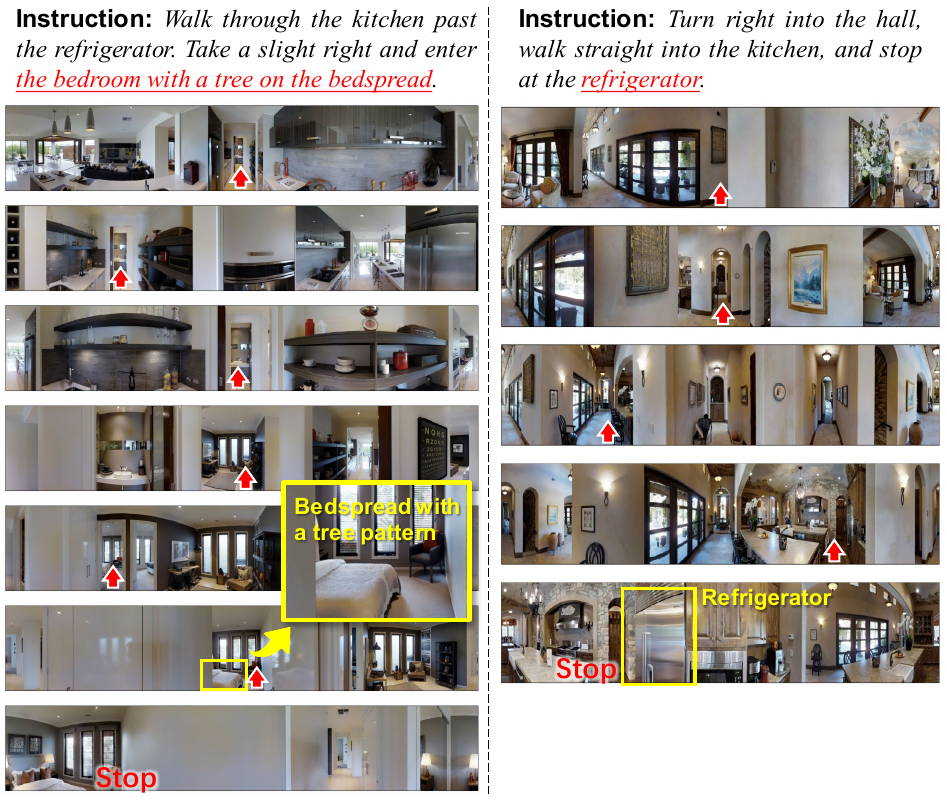}
    \caption{Panoramic paths predicted by MAGIC-L on R2R val-unseen.}
    \label{fig_vis_path_pano}
\end{figure}

Fig.~\ref{fig_vis_paths} visually compares the inference paths of MAGIC-L, MAGIC-S, and GOAT~\cite{wang2024causal} on the R2R validation unseen dataset, highlighting both correct and incorrect predictions. Our qualitative analysis reveals significant improvements in the model trained using our method across three dimensions: 
(I) \textit{Precision in Action Interpretation}: The MAGIC models demonstrate enhanced precision in interpreting directional actions. For instance, as depicted in Fig.~\ref{fig_vis_paths}(a) and (b), the models accurately stop at the bottom of the stairs and turn right instead of left.
(II) \textit{Semantic Alignment and Localization}: Fig.~\ref{fig_vis_paths}(c), (d), and (e) show that MAGICs effectively identify and respond to key semantic references, such as a bedspread with a tree pattern, a closet, and a refrigerator, making correct navigational decisions based on these cues.
(III) \textit{Efficiency in Navigation}: Fig.~\ref{fig_vis_paths} also illustrates the efficiency of navigation. MAGIC-L exhibits the most direct and accurate route predictions, demonstrating superior navigational efficiency. Although MAGIC-S occasionally takes detours, it generally maintains a correct directional course. Additionally, the visualization of egocentric panoramic trajectories for scenarios (c) and (e) in Fig.~\ref{fig_vis_path_pano} further helps to convey the detailed semantics and navigational decisions more clearly.

\subsubsection{Failure Case Analysis}
In addition to visualizing successful cases, we also conducted an in-depth analysis of instances where our models failed. As illustrated in Fig.~\ref{fig_vis_paths}(f), none of the three models were able to accurately predict and follow the given instructions. Upon closer examination of the specific instruction involved, we noticed that it primarily provided a sequence of directional actions without sufficient semantic references as \textit{``step forward, make a left, turn right."} This lack of semantic cues led to confusion in the models' navigation, ultimately directing them towards incorrect destinations. Although both MAGIC-L and MAGIC-S models halted near \textit{``a chair in the room,"} they failed to accurately determine the timing for a right turn, resulting in them entering the wrong room. To mitigate such errors, it is advisable to enrich the semantic content of instructions via some image caption models~\cite{li2022blip}. Integrating the large model's strong decision-making ability~\cite{touvron2023llama} could also help to address such complex situations. We leave this for our future work.

% =============
% 6-Experiment on real
% =============
\section{Sim-to-Real Experiments}
\label{sec_exp_sim2real}

\subsection{Experimental Settings}

\subsubsection{Datasets}
\begin{figure}[t]
    \centering
    \includegraphics[width=\linewidth]{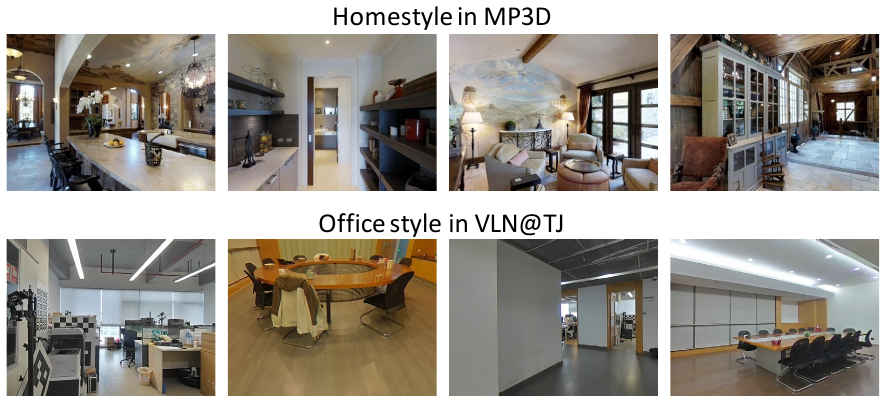}
    \caption{Comparison of MP3D~\cite{chang2017matterport3d} and our collected dataset VLN@TJ.}
    \label{fig_vln_tj}
\end{figure}

To further validate the applicability of our method for enhancing VLN in real-world scenarios with actual robots, we developed a custom-wheeled robot platform and conducted data collection across various campus settings, including offices and conference rooms. The comparison between MP3D scenes and our dataset (\texttt{VLN@TJ}) is provided in Fig.~\ref{fig_vln_tj}. We employed Simultaneous Localization and Mapping (SLAM)~\cite{hess2016real} to create 2D maps and establish a series of discrete navigable points. A panoramic camera was used to capture visual images at these points. Following the simulator's configuration~\cite{anderson2018vision,anderson2021sim}, these panoramic images were segmented into 36 sub-images. The paths are randomly sampled from these discrete topological maps. To ensure diverse linguistic guidance for each path, we engaged three different volunteers to annotate each path with one instruction, resulting in three instructions per path. The dataset we compiled encompasses data from 5 scenes, each containing at least two to three rooms, comprising 136 paths and 408 instructions. This dataset is divided into a training set, which includes 3 scenes with 110 paths, and a test set featuring two previously unseen scenes with 26 paths. The average number of navigable points per scene is 29.8, and the average path length is 9.75 meters. 

\subsubsection{Implementation Details}
We test the navigation performance of DUET~\cite{chen2022think}, GOAT~\cite{wang2024causal}, and our MAGICs. We first load their best checkpoint from R2R and then fine-tune the models on VLN@TJ. All these methods are fine-tuned with a batch size of $8$ and a learning rate of $5\times10^{-6}$ for 80,000 iterations. Interactive continual learning is not applied due to the small size of this self-collected dataset, which could lead to significant overfitting on the test set. After training, we deployed the models onto the mobile robot's processor to evaluate their running speed. The robot is equipped solely with a CPU, specifically an Intel(R) Core(TM) i7-4600U CPU operating at 2.10 GHz. This processor features 2 physical cores and 4 logical processors.

\subsection{Experimental Results}
\begin{figure}[t]
    \centering
    \includegraphics[width=\linewidth]{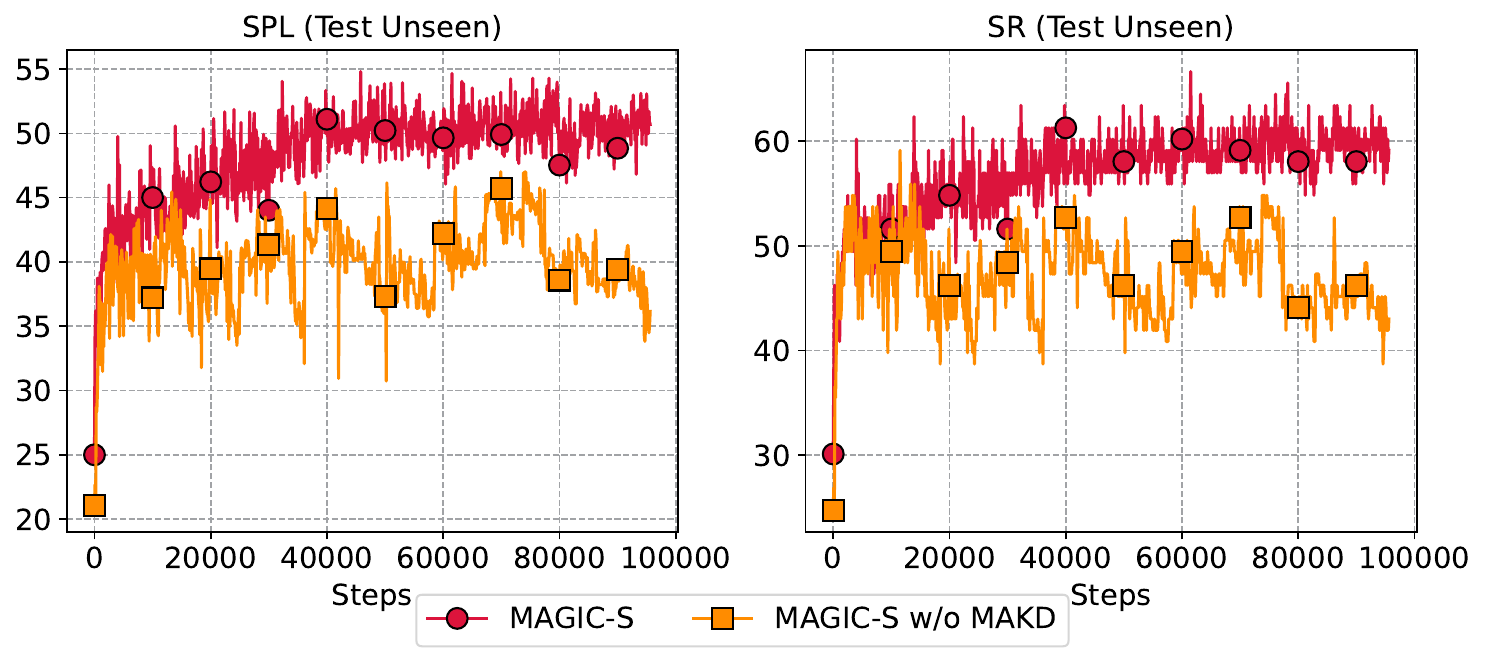}
    \caption{Learning curves of MAGIC-S on the VLN@TJ test unseen set.}
    \label{fig_magics_vlntj}
\end{figure}

Tab.~\ref{tab:sim-to-real} presents the experimental results on the test set of VLN@TJ. Fig.~\ref{fig_magics_vlntj} illustrates the comparisons of learning curves. 
When the MAKD loss term was omitted, the SR of MAGIC-B and MAGIC-S relatively decreased by 15.0\% and 11.3\%, and the SPL dropped by 15.7\% and 19.4\%, respectively. Remarkably, MAGIC-S, despite having the fewest parameters and the fastest operating speed (14 times the FPS of the GOAT), achieved the best generalization performance on this dataset, surpassing GOAT by 13.8\% in SPL. This finding underscores that in real-world scenarios where data availability is limited, our proposed method enables high generalization performance with minimal model complexity, better promoting practical application.

\begin{table}[t]
\centering
\caption{Comparison on the test set of VLN@TJ.}
\label{tab:sim-to-real}
\begin{tabular}{l|cccc|c}
\toprule
Method & SR$\uparrow$ & SPL$\uparrow$ & NE$\downarrow$ & OSR$\uparrow$ & FPS (Hz)$\uparrow$ \\ \midrule
DUET~\cite{chen2022think} & 53.76 & 47.14 & 4.35 & 60.22 & 0.87 \\
GOAT~\cite{wang2024causal} & 65.67 & 48.04 & 3.30 & 75.27 & 1.02 \\ \midrule
MAGIC-B w/o KD & 54.84 & 41.67 & 4.30 & 75.27 & 3.79 \\
MAGIC-B & 64.52 & 49.45 & 3.04 & 76.34 & 3.79 \\
MAGIC-M & 56.99 & 47.64 & 4.40 & 69.89 & 6.49 \\
MAGIC-S w/o KD & 59.14 & 44.04 & 4.08 & 74.19 & 14.75 \\ 
\rowcolor{rowgray}
\textbf{MAGIC-S} & \textbf{66.67} & \textbf{54.66} & \textbf{3.20} & \textbf{78.49} & \textbf{14.75} \\ \bottomrule
\end{tabular}
\end{table}

% =============
% 7-Conclusion
% =============
\section{Conclusion}
\label{sec_conclusion}
This paper aims to improve both the efficiency and effectiveness of the embodied agents in the VLN task. Redundant model parameters often lead to overfitting on small datasets and hinder real-world applications. Therefore, we propose a novel MAGIC method. Specifically, the MAKD is proposed for comprehensive knowledge transferring between the teacher and student models by decoupling and refining the essential meta-abilities. To tackle the potential imbalance in optimizing multiple losses, we propose the MKRW method, which improves network generalization without additional costs. To mitigate the negative effects of an uncertain teacher model, we introduce the MKTD, which dynamically adjusts loss weights at the sample level. Additionally, we suggest the ICoD strategy to improve both teacher and student models iteratively. Experiments are conducted on both the public datasets (R2R and RxR) and our self-built dataset (VLN@TJ). Our largest model, MAGIC-L, achieves new SoTA performance. Our smallest model, MAGIC-S, also outperforms other methods except for the original teacher model under the same training data. In conclusion, our MAGIC method shows great potential in achieving high performance with low model complexity. \textbf{\textit{This, as its name suggests, exhibits a magical effect.}}

% =============
% References
% =============
\bibliographystyle{IEEEtran}
\bibliography{main}

\end{document}